\documentclass[11pt]{article}
\usepackage[final]{acl}
\usepackage{times}
\usepackage{latexsym}
\usepackage[table]{xcolor}
\usepackage[T1]{fontenc}
\usepackage[utf8]{inputenc}
\usepackage{microtype}
\usepackage{inconsolata}
\usepackage{graphicx}
\usepackage{multirow}
\usepackage{amsmath}
\usepackage{amssymb}
\usepackage{booktabs}
\usepackage{subcaption}
\usepackage{bm}
\usepackage{enumitem}
\usepackage{makecell}
\usepackage{pifont}
\usepackage{longtable}
\usepackage{array}

\usepackage[absolute,overlay]{textpos}
\usepackage{graphicx}
\usepackage{tcolorbox}
\tcbuselibrary{breakable, skins, external} 

\definecolor{bg_gray}{RGB}{245,245,245}
\definecolor{frame_gray}{RGB}{220,220,220}
\definecolor{meta_color}{RGB}{100,100,100}

\newtcolorbox{benchbox}[1][]{
  enhanced jigsaw, 
  breakable, 
  colback=bg_gray,
  colframe=frame_gray,
  arc=3mm,
  boxrule=1pt,
  title={\textbf{#1}},
  coltitle=black,
  fonttitle=\large\bfseries,
  left=6mm,
  right=3mm,
  top=2mm, 
  bottom=3mm,
  fontupper=\small\ttfamily,
  before skip=10pt,
  after skip=10pt
}

\newcommand{\qtype}[1]{\noindent\textbf{\color{blue!40!black}#1}}
\newcommand{\timestamp}[1]{\hfill{\small\ttfamily\color{meta_color}[#1]}}
\newcommand{\lightsep}{
  \par\vspace{-0.5em}
  \noindent{\color{gray!30}\hrule height 0.5pt}
  \vspace{0.5em}\par
}
\newcommand{\cmark}{\checkmark}
\newcommand{\xmark}{\ding{55}}

\usepackage{cleveref}

\title{Evaluating Memory Capability in Continuous Lifelog Scenario}

\author{ 
  \textbf{Jianjie Zheng}$^{1,2}$\thanks{~~Equal contribution. Work done during an internship at RayNeo AI.}, 
  \textbf{Zhichen Liu}$^{1,2\ast}$,  
  \textbf{Zhanyu Shen}$^{2,4}$,  
  \textbf{Jingxiang Qu}$^{2,5}$ \\ 
  \textbf{Guanhua Chen}$^{1\dagger}$,  
  \textbf{Yile Wang}$^4$,  
  \textbf{Yang Xu}$^1$,  
  \textbf{Yang Liu}$^3$,  
  \textbf{Sijie Cheng}$^{2,3}$\thanks{~~Corresponding authors.} \\ 
  \addlinespace[0.1cm] 
  $^1$Southern University of Science and Technology, \quad $^2$RayNeo.AI \\ 
  $^3$Tsinghua University, \quad $^4$Shenzhen University, \quad $^5$Shanghai Jiao Tong University
}

\begin{document}

\begin{textblock*}{5cm}(1.5cm,1cm) 
    \includegraphics[width=3cm]{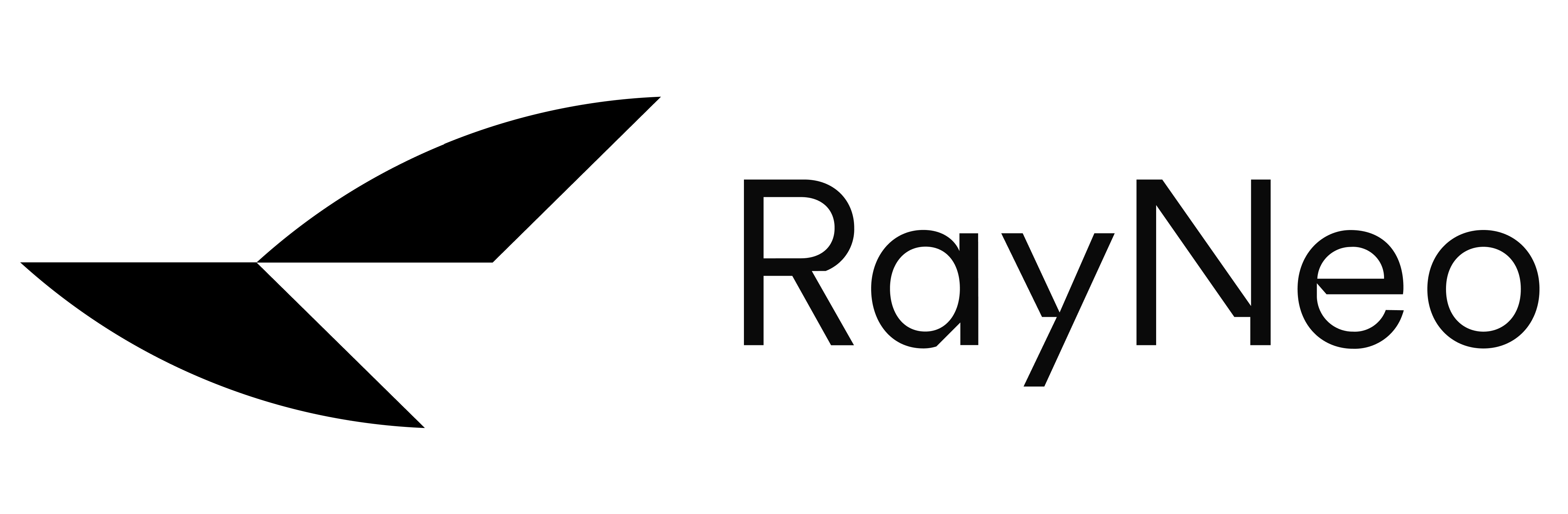}
\end{textblock*}

\maketitle

\begin{abstract}
Nowadays, wearable devices can continuously lifelog ambient conversations, creating substantial opportunities for memory systems.
However, existing benchmarks primarily focus on online one-on-one chatting or human-AI interactions, thus neglecting the unique demands of real-world scenarios.
Given the scarcity of public lifelogging audio datasets, we propose a hierarchical synthesis framework to curate \textbf{\textsc{LifeDialBench}}, a novel benchmark comprising two complementary subsets: \textbf{EgoMem}, built on real-world egocentric videos, and \textbf{LifeMem}, constructed using simulated virtual community.
Crucially, to address the issue of temporal leakage in traditional offline settings, we propose an \textbf{Online Evaluation} protocol that strictly adheres to temporal causality, ensuring systems are evaluated in a realistic streaming fashion.
Our experimental results reveal a counterintuitive finding: current sophisticated memory systems fail to outperform a simple RAG-based baseline. This highlights the detrimental impact of over-designed structures and lossy compression in current approaches, emphasizing the necessity of high-fidelity context preservation for lifelog scenarios.
\end{abstract}

\section{Introduction}

Large language models (LLMs) have demonstrated remarkable capabilities across a wide range of tasks~\citep{openai2022chatgpt, gpt4, qwen3}, especially in the single-turn scenario with short-term conversational context.
Subsequently, LLMs show superior reasoning ability as automatic agents to process a series of complex tasks in real world~\citep{toolformer,gpt4tools}, meanwhile, place a higher requirement on the context length.
To explore long-term memory capability of LLMs, one line of works~\citep{longlora,llama3,qwen3} focus on probing the accuracy of locating evidence in extremely long-context passages, such as Needle In A Haystack~\citep{naih}.
However, the strategy of increasing context length indefinitely is not a solution to long-term memory, due to the exponential growth in inference costs and the ability of long-term memory utilization~\citep{ruler, loogle, lost-in-the-middle}.
Consequently, the development of memory system has emerged. It requires LLMs to adaptively remember and retrieve relative evidence from massive information over extended periods.

\begin{figure}[!t]
    \centering
    \includegraphics[width=\linewidth]{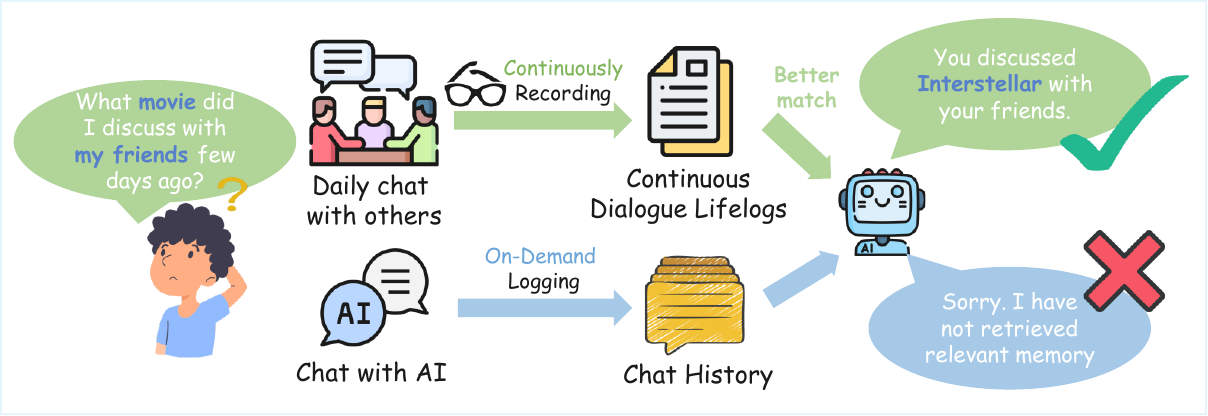}
    \caption{Comparison between (1) The microphone-always-on scenario, which continuously recording dialogue with others in daily life, and (2) Chatting with AI scenario, which on-demand logging to form the chat history.}
    \label{fig:scenario}
    \vspace{-1.5em}
\end{figure}

\begin{figure*}[!t]
    \centering
    \includegraphics[width=\linewidth]{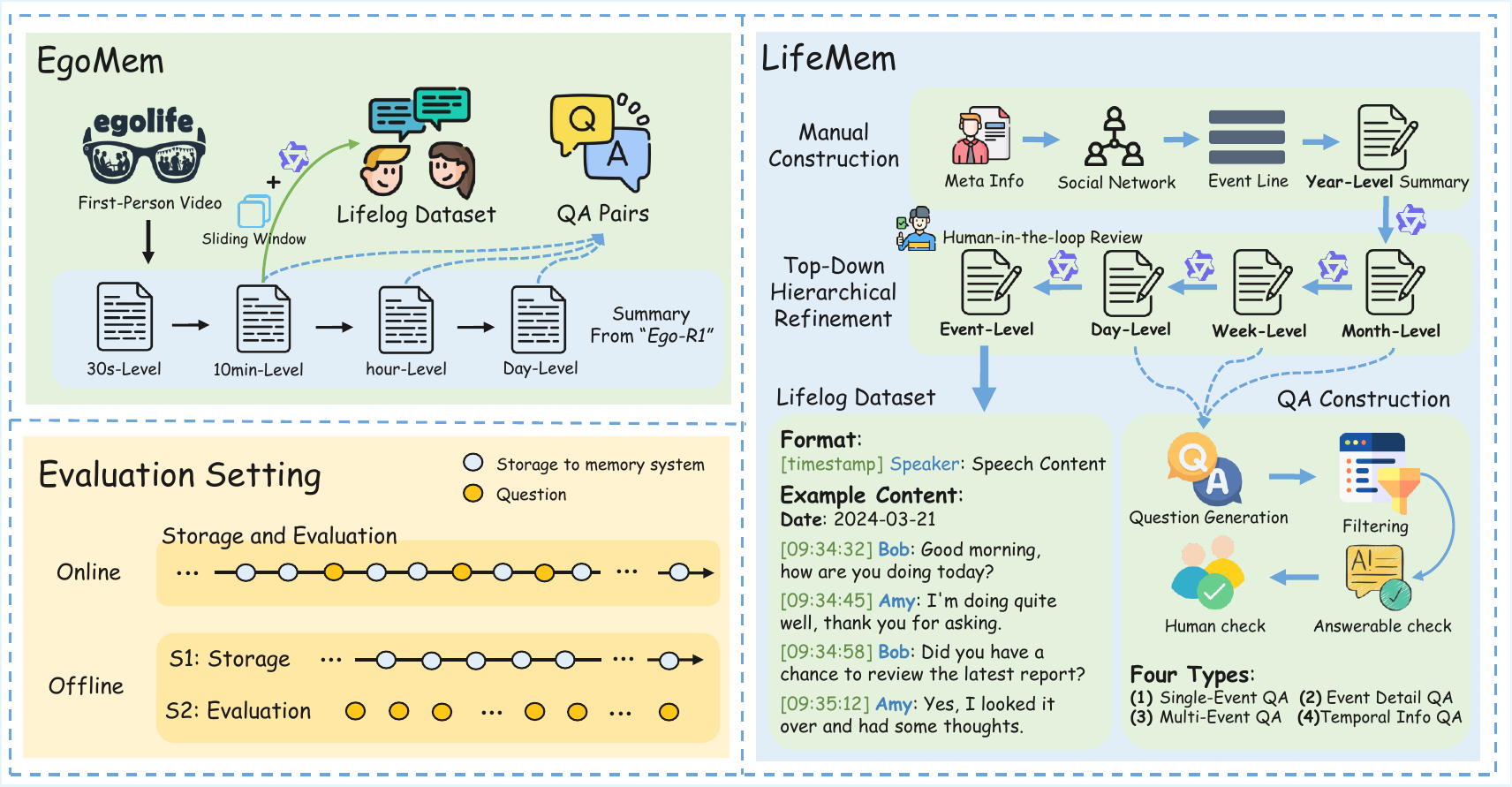}
    \vspace{-10pt}
    \caption{We introduce \textsc{LifeDialBench}, which consists of two subsets: EgoMem (top left), constructed from real-world egocentric videos (EgoLife), and LifeMem (right), a more comprehensive dataset built upon a Human-in-the-loop Hierarchical Life-Simulation Framework. Additionally, we propose a novel online evaluation method that assesses model performance incrementally during data storage, in contrast to conventional evaluation conducted only at the end of storage.}
    \label{fig:main}
    \vspace{-10pt}
\end{figure*}

Meanwhile, there exist various benchmarks primarily focus on dialogue scenarios, covering Person-AI interaction~\citep{personamem,longmemeval} and dyadic dialogues (i.e., person-person conversations;~\citet{locomo}).
However, the above-mentioned studies neglect a promising scenario as illustrated in~\Cref{fig:scenario}: \textbf{\textit{continuous dialogue lifelogs}}. 
Nowadays, there emerges a series of commercial wearable devices with potential to achieve microphone always-on, such as smart glasses (e.g., \href{https://www.ray-ban.com/usa/ray-ban-meta-ai-glasses}{Ray-Ban Meta}, \href{https://rayneo.cn/}{RayNeo V3/X3}, \href{https://www.mi.com/prod/xiaomi-ai-glasses}{Xiaomi AI Glasses}), and recording machines (e.g., \href{https://www.plaud.ai/?srsltid=AfmBOoqeWm89RQ1FfFz9VWWNNve2ATsM0PcelPsDVGxaZRibXGk4rLAo}{Plaud}).
Equipped with these wearable devices, users can continuously record the surrounding audio which fully filled with intensive dialogue content.
Using automatic speech recognition (ASR), the audio stream is transcribed into text and stored in a long-memory database after post-processing.
Compared to prior passages and Person-AI interaction, continuous dialogue lifelogs have several unique characteristics:
(1) The daily conversations integrate multi-person interactions, casual and temporal event threads, and simulated social networks.
(2) Through round-the-clock recording, the lifelogs enables the AI assistant to accumulate an extensive understanding of users' facts, perfectly embodying the highly promising usage scenario of a personalized assistant.

To systematically evaluate the long-term memory capacity of agents in continuous dialogue lifelogs, we introduce \textbf{\textsc{LifeDialBench}}, which comprises two complementary subsets as shown in Figure~\ref{fig:main}: \textbf{EgoMem} and \textbf{LifeMem}.
Both subsets adopt a hierarchical life simulation framework to generate the dataset.
EgoMem is constructed using a bottom-up (i.e., from second to week) summarization based on the real-life first-person video dataset EgoLife~\citep{EgoLife_Yang_2025_CVPR}, which records egocentric video from six individuals over a seven-day period.
To extend the temporal span and ensure long-horizon coherence, we further use LLMs with a top-down (i.e., from year to day) elaboration to simulate a year-long personal lifelog rich in multi-party conversations, forming the LifeMem.
For both subsets, we generate QA pairs from multi-level event summaries, enabling systematic probing of memory retrieval across different temporal granularities.
Notably, we first propose an \textbf{online evaluation} protocol that follows the linear flow of time with information updates and conflicts, offering a more realistic assessment of long-term memory in real-world conditions.

We evaluate four representative memory systems on \textsc{LifeDialBench}.
Unexpectedly, a simple RAG baseline consistently outperforms specialized methods, highlighting the advantage of raw text preservation over the lossy compression inherent in current designs.
Furthermore, we identify temporal retrieval as a universal bottleneck across all methods.
Crucially, our analysis validates the necessity of online evaluation, demonstrating that traditional offline settings distort assessment by permitting temporal leakage and misjudging dynamic memory updates.
These findings collectively underscore the dual importance of context fidelity and strict temporal linearity in next-generation lifelog systems.
\footnote{We release our code and data at \url{https://github.com/RayNeo-AI-2025/LifeDialBench}.}

\begin{table*}[tbp]
\centering
\caption{
Comparison of memory benchmarks. 
The key properties are summarized including the type of scenario (\textbf{Scenario}), the temporal coverage (\textbf{Time Span}), number of sessions (\textbf{\#Sessions}), whether continuous recording is supported (\textbf{Cont. Rec.}), whether the queries contain explicit timestamp for simulation (\textbf{TS}), whether support for online evaluation (\textbf{Online}). 
}
\label{tab:benchmark comparison}
\resizebox{1\linewidth}{!}{
\begin{tabular}{lccccccc}
\toprule
Benchmark     & Scenario          & Time Span      & \#Sessions & Cont. Rec. & TS   & Online \\ 
\midrule
LoCoMo \citep{locomo}        & Person-Person    & Few months     & 1k         & \xmark     & \xmark & \xmark \\ 
MemoryBank \citep{memorybank}    & Person-AI       & 10 days        & 300        & \xmark     & \cmark & \xmark \\ 
LongMemEval \citep{longmemeval}   & Person-AI       & N/A            & 50k        & \xmark     & \cmark & \xmark \\ 
MemBench \citep{membench_gaoling}      & Person-AI       & N/A            & 65k        & \xmark     & \xmark & \xmark \\ 
\midrule
\textbf{EgoMem (Ours)}   & Multi-Person    & 7 days         & 1.7k       & \cmark     & \cmark & \cmark \\ 
\textbf{LifeMem (Ours)}  & Multi-Person   & 1 year         & 3.8k       & \cmark     & \cmark & \cmark \\ 
\bottomrule
\end{tabular}}
\end{table*}

\section{Related Work}

\paragraph{Memory Systems.}
Modern long-term memory systems for large language models (LLMs) typically adopt a modular architecture to handle extended contexts. These systems generally consist of a memory manager for database maintenance, a summary agent for information compression~\citep{amem}, and a retriever for context-aware fetching. 
Various implementations have been explored to optimize this framework. 
One prevailing approach involves using a Summary Agent to condense historical interactions into concise insights before storage in vector databases~\citep{scm, amem, mem0}. 
To better capture structural dependencies, recent works such as \citet{mem0}, \citet{hipporag}, and \citet{zep} have transitioned from flat vector embeddings to graph-based representations. 
Furthermore, inspired by traditional computing, works like \citet{memgpt} and \citet{memos} treat LLM context as a tiered memory hierarchy, managing data through mechanisms analogous to operating systems. 
A systematic categorization of these architectural paradigms is provided in \Cref{fig:memory_system} in the Appendix.

\paragraph{Benchmarks for Long-Term Memory.}
The rapid evolution of memory systems has spurred the development of specialized evaluation benchmarks, as summarized in \Cref{tab:benchmark comparison}. 
Early efforts like MemoryBank~\citep{memorybank} utilized manually constructed QA pairs for basic retrieval testing. 
Subsequent benchmarks increased complexity by focusing on session-based dynamics; for instance, LoCoMo~\citep{locomo} evaluates person-person dialogues across multiple dimensions, while PersonaMem~\citep{personamem} and LongMemEval~\citep{longmemeval} scale the context up to 1.5M tokens for person-AI interactions. 
They have made significant progress for chatbot-like memory system. However, remaining a gap between real world scenarios--the scenarios of multi-person communication, and the situation where the memory system is continuously activated, such as an always-on personal agent.
In this paper, we propose a benchmark contains multi-person dialogue, which is rolling from day to night, and continuous for year-long. This benchmark behaves closer to the real world compared to previous works.

\section{Benchmarking Lifelog Memory}
To explore the long-term memory capacity of agents in continuous dialogue lifelogs, we specifically construct a benchmark named \textbf{\textsc{LifeDialBench}} with two complementary subsets for egocentric memory\citep{cheng2024videgothink}, as illustrated in \Cref{fig:scenario}.
The first subset, named \textbf{EgoMem}, is constructed based on the existing EgoLife dataset~\citep{EgoLife_Yang_2025_CVPR} which contains daily video recording across seven days.
Moreover, to mimic the continuous dialogue lifelogs in real-life with more time span and scene diversity, we further adopt data synthesis to construct a year-long subset, named \textbf{LifeMem}.
Both subsets are constructed with a hierarchical life simulation framework, where we use bottom-up and top-down manners due to different data sources. Specifically, all dialogue and summary content are generated using \texttt{Qwen3-235B-Instruct}\footnote{Corresponds to  \href{https://huggingface.co/Qwen/Qwen3-235B-A22B-Instruct-2507}{\texttt{Qwen3-235B-A22B-Instruct-2507}}}.

More details will be introduced in this section.

\subsection{Design Principles}

Constructing a rigorous benchmark for lifelog memory evaluation requires confronting
challenges that are qualitatively distinct from those in standard dialogue settings.
We identify three core principles that govern the design of \textsc{LifeDialBench}.

\paragraph{Temporal Causality.}

In realistic lifelogging, an agent must respond to a query using \emph{only} information
available up to that moment---future context is physically inaccessible.
Traditional offline evaluation violates this constraint by granting agents access to the
complete dataset prior to answering any query, introducing \textit{temporal leakage} that
systematically overestimates real-world performance.
This motivates a causally-constrained \textbf{Online Evaluation Protocol}, formally
defined in \S\ref{sec:online_eval}, which enforces a streaming interaction flow strictly
aligned with the physical arrow of time.

\paragraph{Compositional Query Complexity.}

Lifelog queries extend well beyond simple fact retrieval.
They demand (i)~\textit{Temporal Grounding}---localizing \emph{when} an event occurred
relative to the query moment---and (ii)~\textit{Long-Horizon Multi-Hop Reasoning}---synthesizing disjoint events dispersed across days or months to form a coherent answer. Benchmarks built on isolated dialogue sessions fail to stress-test these capabilities.
Accordingly, \textsc{LifeDialBench} incorporates four structured question types
(\texttt{single\_event}, \texttt{multi\_event}, \texttt{time\_query}, \texttt{event\_detail})
to systematically probe both dimensions, as detailed in \S\ref{sec:qa_construction}.

\paragraph{Ecological Validity of the Lifelog Stream.}

Authentic lifelogging differs from mere session concatenation along two critical axes.
First, the data stream is \textit{multi-party}: interactions span diverse social
contexts---family members, colleagues, and strangers---rather than a fixed dyadic pair,
introducing speaker heterogeneity absent from prior benchmarks.
Second, the timeline is \textit{semantically coherent}: topics evolve, recur, and
interleave organically across time rather than resetting at session boundaries.
Together, these properties demand data that preserves genuine chronological continuity and multi-speaker dynamics, which directly guides the hierarchical construction methodology described as follows.

\subsection{EgoMem}\label{sec:EgoMemBench}

\paragraph{Data Source.}
We construct EgoMem based on the Ego-R1 corpus~\citep{egor1}, which provides multi-scale textual summaries of the real-world egocentric dataset EgoLife~\citep{EgoLife_Yang_2025_CVPR}. Initially, we attempted to transcribe EgoLife's raw audio via Automatic Speech Recognition (ASR). However, the inherent sparsity of continuous conversation and severe ASR degradation in noisy daily environments made it impractical to extract coherent transcripts necessary for constructing high-quality Q\&A pairs. To overcome this, we propose a hybrid proxy approach: we leverage Ego-R1's structured 10-minute summaries as factual anchors to prompt an LLM to synthesize lifelog-style dialogues. This strategy produces \textit{summary-grounded simulated dialogues} that preserve the authentic physical event sequences of the real world while yielding contextually rich conversational streams suitable for rigorous memory evaluation.

\paragraph{Data Curation.}
To maintain narrative coherence across extended temporal horizons, we employ a \textit{sliding-window generation strategy} at a 10-minute granularity. Rather than generating isolated dialogue segments, the LLM is conditioned on a dynamic historical context window (the preceding 60 minutes) alongside the target summary. This empirically determined window balances speaker consistency, relational continuity, and topic flow while mitigating hallucinations and repetitive patterns.

\paragraph{Quality Assurance \& Conversational Grounding.}
To ensure our synthesized data reflects the natural dynamics of daily communication rather than overly formalized written prose, we implement a human-LLM collaborative review pipeline~\citep{text-grad}. Beyond standard \textit{Factual Consistency} checks against the source summaries, we specifically optimize for \textit{Conversational Naturalness}. Human annotators utilize LLM-flagged feedback to iteratively revise the text, ensuring the dialogues capture casual interactions, appropriate tone, and implicit contexts typical of real-world continuous recordings. Comprehensive details regarding ASR limitations, granularity selection, and the full review workflow are provided in Appendix~\ref{app:egomem_details}.

\subsection{LifeMem}

While EgoMem provides grounded real-life recordings, it is constrained by a seven-day window and limited social diversity. To overcome these barriers and enable the study of long-term memory at scale, we develop a \textbf{Human-in-the-loop (HITL) Hierarchical Life-Simulation Framework} to synthesize LifeMem—a year-long continuous dialogue lifelog. Diverging from conventional end-to-end synthesis, our framework treats lifelog synthesis as a principled, multi-stage trajectory expansion process where humans steer the narrative consistency and logical grounding.

\paragraph{Hierarchical Synthesis Framework.} 
The core of LifeMem is a top-down refinement strategy that decomposes a person's entire year into granular, dialogue-centric records across four stages: 

\noindent\textbf{1) Identity and Social Seeding:} We first manually define a virtual agent's persona (e.g., age, occupation, personality) and a comprehensive social network encompassing family, colleagues, and friends. Human curators serve as the system architects, validating the realism of these relationships to ensure a stable social foundation for long-term interactions. This social network serves as the basis for generating multi-party conversations that reflect realistic interpersonal dynamics.

\noindent\textbf{2) Multi-dimensional Event Trajectories:} Rather than generating random experiences, we construct eleven distinct \textit{event lines} (e.g., professional growth, health, household management) across key life dimensions. These event lines serve as structured narrative threads that capture diverse and realistic life dynamics, establishing a solid basis for generating lifelog data that maintains semantic richness and long-horizon coherence. We then align all event lines into a \textit{year-level summary}, which serves as the backbone of the year, ensuring that an event mentioned in January (e.g., a project kickoff) has logical echoes in subsequent months.

\noindent\textbf{3) Iterative Refinement (Allocation \& Enrichment):} The framework progressively expands high-level annual summaries into monthly, weekly, and daily scales through a top-down refinement strategy. This involves two sub-steps: (i) \textit{Allocation:} LLMs distribute annual narrative into monthly and weekly placeholders. (ii) \textit{Enrichment:} Each placeholder is expanded into detailed narratives while preserving higher-level context. To further align the simulated lifelog with real-world temporal structures, we incorporate external calendar signals such as statutory holidays, weekends, and workdays. Notably, directly generating lifelogs from daily experience often leads to coarse or repetitive descriptions, as LLMs struggle to capture the fine-grained variations that naturally arise within a day; our two-stage refinement effectively mitigates this issue.

\noindent\textbf{4) Dialogue Grounding:} Daily experiences are further segmented into \textit{event-level narratives}, each annotated with temporal boundaries, locations, and participants. Finally, we synthesize continuous dialogue streams conditioned on these fine-grained event contexts, the virtual user's background, and the social relationship network, ensuring natural conversational flow and long-horizon coherence.

\paragraph{Human-in-the-Loop Calibration.} 

Crucially, our framework operates as a transparent and controllable pipeline; it integrates human intervention at every hierarchical transition to mitigate the ``stochastic parrot'' effect and temporal drift. As detailed in the calibrated workflow (\Cref{app:human-in-the-loop}):

\begin{itemize}[leftmargin=*,itemsep=0pt]
    \item \textbf{Consistency Auditing:} After each refinement stage (e.g., Year-to-Month), human annotators audit the generated trajectories. They identify and correct temporal contradictions (e.g., a character being in two places at once) or identity inconsistencies.
    \item \textbf{Manual Trajectory Pruning:} Annotators prune repetitive or mundane event patterns that frequently emerge in vanilla LLM outputs, ensuring the lifelog maintains high information density and narrative diversity.
    \item \textbf{Quality Gatekeeping:} Data only advances to the next level of granularity (e.g., Week-to-Day) once it passes a human-verified quality check. This prevents error propagation, ensuring that the final dialogues are grounded in a logically sound and human-validated annual history.
\end{itemize}

By combining automated hierarchical expansion with rigorous human steering, LifeMem achieves a balance of scalability and high-fidelity realism, providing a robust foundation for benchmarking memory systems in privacy-preserving, always-on scenarios.

\subsection{Question-Answering Pairs Curation}
\label{sec:qa_construction}

\paragraph{Task Formats.}
\textsc{LifeDialBench} supports two complementary question-answering formats: \textit{Multiple-Choice Questions (MCQ)} and \textit{Open-Ended QA}. MCQ strictly assesses precise retrieval capacity, while Open-Ended QA—where reference answers are derived from the correct MCQ options—evaluates the agent's generative synthesis capability via LLM-as-a-Judge~\citep{llm-as-a-judge}. 

\paragraph{Question Types.}\label{sec:question_types}
To comprehensively evaluate memory retrieval and reasoning~\citep{locomo}, we design four distinct question types (consistent with labels in \Cref{fig:main}):
(i) \textit{QT1: Single-Event QA}: Retrieving core event content based on ambiguous queries.
(ii) \textit{QT2: Event Detail QA}: Pinpointing specific snippet-level event attributes.
(iii) \textit{QT3: Multi-Event QA}: Retrieving and integrating information across multiple disconnected events over time.
(iv) \textit{QT4: Temporal Info QA}: A lifelog-specific type requiring the exact timestamp or temporal relation of a given event.

\paragraph{Question-Answering Construction.}
We prompt \texttt{Qwen3-235B-Thinking}
\footnote{Corresponds to \href{https://huggingface.co/Qwen/Qwen3-235B-A22B-Thinking-2507}{\texttt{Qwen3-235B-A22B-Thinking-2507}}}
~\citep{qwen3} to synthesize QA pairs spanning multiple temporal granularities. To ensure rigorous evaluation, we implement a strict quality assurance pipeline. 

\noindent\textbf{1) Causality Preservation:} All distractors are generated exclusively using events prior to the query timestamp, preventing agents from trivially eliminating ``future'' options during online streaming. 

\noindent\textbf{2) Leakage Filtering:} A secondary \texttt{Qwen3-32B}
rewrites questions to eliminate explicit timestamp leakages, enforcing semantic understanding over temporal shortcuts. 

\noindent\textbf{3) Answerability Check:} We prompt \texttt{qwen3-max}
\footnote{Corresponds to  \href{https://bailian.console.aliyun.com/cn-beijing?spm=5176.29619931.nav-v2-dropdown-menu-0.d_main_2_0.74cd10d7sBHNbm&tab=model&scm=20140722.M_10944401._.V_1\#/model-market/detail/qwen3-max-2025-09-23}{\texttt{qwen3-max-2025-09-23}}}
to answer the generated questions. To mitigate random guessing (a 25\% chance in MCQ), the evaluator must output explicit evidence rationales before selecting an option. Unanswerable questions are discarded, and correct ones undergo human random sampling to mitigate LLM self-preference bias. 

\Cref{tab:answerable-check} details the filtering statistics and near-perfect evaluator accuracy for LifeMem, confirming the genuine answerability of the curated queries. Ultimately, the benchmark retains 1,774 questions for EgoMem and 1,717 for LifeMem. Detailed prompts are in Appendix~\ref{app:question generation prompt}.

\begin{table}[t]
\centering 
    \captionof{table}{The number of generated questions, the number of questions remained after filtering and human annotation, and the model accuracy in the final answerable verification.}
        \resizebox{\linewidth}{!}{
        \begin{tabular}{l|c|c|c|c}
            \toprule
            LifeMem& Total & Filtered & Keep Rate & Model Acc. \\
            \midrule
            Daily   & 1464  & 1430   & 97.68\%    & 99.23\%    \\
            Weekly  & 248   & 241    & 97.18\%    & 98.76\%    \\
            Monthly & 48    & 46     & 95.83\%    & 100\%      \\
            \midrule
            All     & 1760  & 1717   & 97.56\%    & 99.18\%    \\
            \bottomrule
        \end{tabular}
        }
    \label{tab:answerable-check}
\end{table}

\begin{figure*}[t!]
    \centering
    \includegraphics[width=\linewidth]{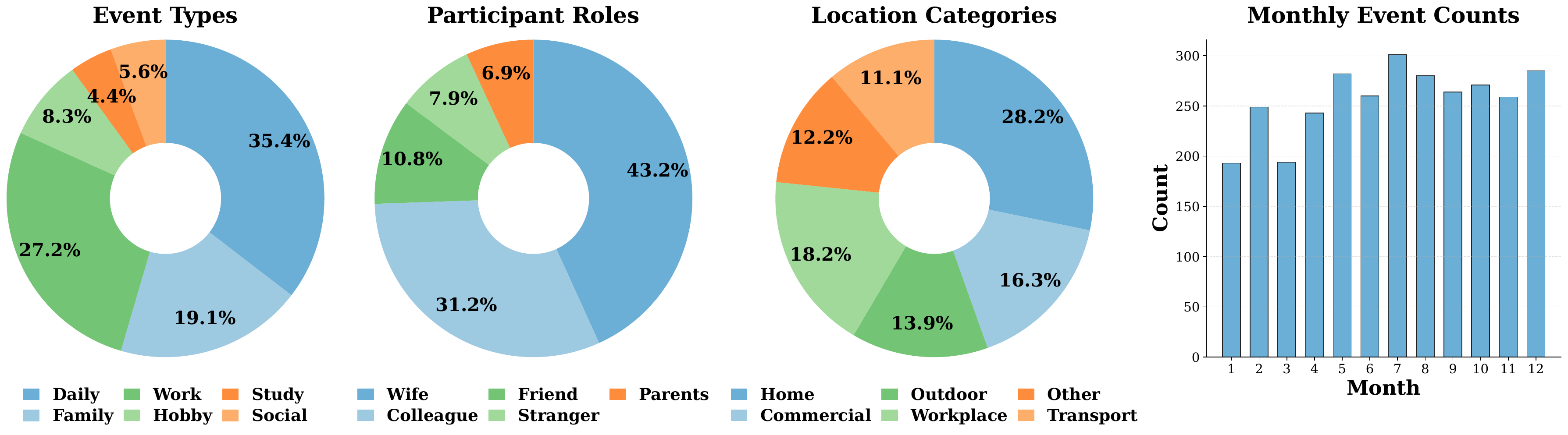}
    \caption{Distributional statistics of the LifeMem dataset. The plots summarize event types, social roles, locations, and monthly dialogue counts, showing that the dataset is balanced and closely aligned with real-world lifelog patterns.}
    \label{fig:LifeMem_dis}    
\end{figure*}

\subsection{Dataset Statistics}

Figure~\ref{fig:LifeMem_dis} illustrates the distributional characteristics of LifeMem across four dimensions: event types, social roles, locations, and monthly dialogue counts.
The dataset covers both routine and higher-level life pursuits, spanning intimate, professional, and casual social interactions across diverse geographical settings.
Crucially, monthly event counts remain stable throughout the year with no significant seasonal bias, which prevents evaluation results from being skewed toward time-specific patterns—a critical property for long-horizon memory benchmarking.

For QA pairs, \Cref{fig:category distribution} in~\cref{app:questino types distribution} shows the proportion of each question type after filtering.
The four types—event content recall, event detail retrieval, multi-hop event reasoning, and temporal information QA—are distributed as 25.3/25.0/25.0/24.6 in LifeMem and 25.1/25.1/24.9/24.9 in EgoMem, ensuring a balanced evaluation of memory systems across multifaceted retrieval and reasoning capabilities.

\subsection{Online Evaluation Protocol}
\label{sec:online_eval}

To rigorously assess memory systems in a realistic ``always-on'' setting, we propose a strict \textbf{Online Evaluation Protocol}. Unlike traditional offline paradigms~\citep{locomo, longmemeval} that batch-process queries after indexing the entire dataset---thereby risking temporal leakage---our protocol enforces a linear, streaming interaction flow that strictly adheres to the physical arrow of time.

Formally, we model the lifelog as a time-ordered stream of data chunks $\mathcal{D} = \{d_1, d_2, \dots, d_T\}$, where each $d_t$ corresponds to a discrete time window (e.g., a dialogue session or a fixed-length segment). A set of queries $\mathcal{Q}_t$ is associated with each time step $t$, representing questions that become answerable only after observing $d_t$. As illustrated in \Cref{fig:main}, the evaluation proceeds as a recursive sequential process:

\noindent\textbf{1) Streaming Ingestion:} At step $t$, the memory system receives and processes the new data chunk $d_t$. The system updates its internal memory state from $\mathcal{M}_{t-1}$ to $\mathcal{M}_t$ based on its specific retention and consolidation mechanisms (e.g., indexing, summarization, or graph updates).

\noindent\textbf{2) State Freezing:} Before revealing any future data $d_{t+1}$, the memory state $\mathcal{M}_t$ is effectively ``frozen.'' This ensures a read-only evaluation snapshot where the system strictly cannot access future information, thereby eliminating look-ahead bias.

\noindent\textbf{3) In-Stream Assessment:} The system is tasked to answer all queries $q \in \mathcal{Q}_t$ using strictly the information available in the frozen state $\mathcal{M}_t$. The performance is recorded for this specific timestamp.

\noindent\textbf{4) Causal Progression:} Only after all queries in $\mathcal{Q}_t$ are resolved does the system advance to time $t+1$, repeating the cycle.

This protocol guarantees \textbf{Temporal Causality}: the answer $a_t$ to a query $q \in \mathcal{Q}_t$ depends solely on the available history $\mathcal{H}_t = \{d_1, \dots, d_t\}$. The necessity of this online mode stems from two fundamental flaws in traditional offline evaluation for lifelog scenarios:

\begin{itemize}[leftmargin=*,itemsep=0pt]
    \item \textbf{Future Context Contamination:} 
    In a causal online setting, the response to query $q_t$ strictly relies on the current memory: $a_t^{\text{real}} = A(q_t \mid \mathcal{M}_t)$. Conversely, offline evaluation grants the system access to the complete history $\mathcal{M}_T$ (where $T$ is the end of the stream), yielding $a_t^{\text{offline}} = A(q_t \mid \mathcal{M}_T)$. Whenever $t < T$, future data ($d_{t+1}, \dots, d_T$) can illicitly influence the response through advanced indexing or global summarization, creating an uncontrolled confound and overestimating the system's promptness.

    \item \textbf{Irreversible Memory Modification:} 
    In many memory systems (especially those with overwrite or consolidation mechanisms), memories evolve dynamically: $\mathcal{M}_t = \text{Update}(\mathcal{M}_{t-1}, d_t)$. In the online mode, $d_t$ is fresh and explicitly represented ($d_t \subseteq \mathcal{M}_t$). However, in offline mode, if $d_t$ is compressed or purged by subsequent updates ($\exists \tau \in (t, T]$ causing the information in $d_t$ to be lost in $\mathcal{M}_T$), the system will fail to answer $q_t$ retrospectively. Thus, offline metrics fail to capture the system's ability to provide real-time responses before information decay occurs, severely misestimating a model's real-world dynamic retrieval capacity.
\end{itemize}

\begin{table*}[htbp]
\centering
\vspace{-20pt}
\setlength{\tabcolsep}{4.5pt} 
\renewcommand{\arraystretch}{1} 
\caption{Main results of memory systems' performance on \textsc{LifeDialBench} in the online evaluation setting. The method with the best overall performance are \textbf{bold}, the second are \underline{underlined}.}
\label{tab:main_results}
\resizebox{\linewidth}{!}{
\begin{tabular}{c|c|ccccc|ccccc}
\toprule
 & & \multicolumn{5}{c|}{Open-Ended} & \multicolumn{5}{c}{Multiple-Choices} \\
Models & Method & QT1 & QT2 & QT3 & QT4 & Overall & QT1 & QT2 & QT3 & QT4 & Overall \\
\midrule
\multicolumn{12}{c}{\textbf{EgoMem}} \\
\midrule
\multirow{4}{*}{gpt-4o-mini} 
 & RAG   & 39.11 & 57.91 & 10.91 & 26.57 & \textbf{34.07} & 73.79 & 86.25 & 60.69 & 57.20 & \textbf{69.86} \\
 & A-Mem & 36.29 & 53.33 & 8.73  & 30.18 & \underline{32.48} & 70.56 & 86.66 & 52.83 & 55.40 & \underline{66.77} \\
 & Mem0  & 15.72 & 17.91 & 3.49  & 16.21 & 13.41          & 56.85 & 53.75 & 44.54 & 37.83 & 48.56 \\
 & MemOS & 26.61 & 33.75 & 6.55  & 11.71 & 20.02          & 69.35 & 72.91 & 66.37 & 38.73 & 62.30 \\
\midrule
\multirow{4}{*}{qwen-plus} 
 & RAG   & 37.50 & 59.16 & 13.10 & 31.08 & \underline{35.56} & 70.16 & 85.41 & 60.26 & 56.30 & \textbf{68.37} \\
 & A-Mem & 43.54 & 58.75 & 14.41 & 33.33 & \textbf{37.91}    & 71.37 & 86.66 & 55.89 & 55.40 & \underline{67.73} \\
 & Mem0  & 12.09 & 16.66 & 6.11  & 13.96 & 12.24             & 45.96 & 57.91 & 36.68 & 35.13 & 44.19 \\
 & MemOS & 33.46 & 44.58 & 17.46 & 35.58 & 32.90             & 64.91 & 76.25 & 58.51 & 55.40 & 64.00 \\
\midrule
\multicolumn{12}{c}{\textbf{LifeMem}} \\
\midrule
\multirow{4}{*}{gpt-4o-mini} 
 & RAG   & 36.86 & 73.25 & 20.41 & 23.12 & \textbf{38.46} & 74.42 & 92.09 & 69.93 & 56.76 & \textbf{73.63} \\
 & A-Mem & 33.63 & 74.44 & 19.71 & 21.72 & \underline{37.40} & 72.34 & 90.42 & 69.69 & 54.19 & \underline{71.98} \\
 & Mem0  & 7.36  & 21.47 & 5.39  & 8.64  & 10.72          & 67.96 & 69.90 & 68.06 & 36.67 & 60.97 \\
 & MemOS & 23.21 & 56.87 & 11.39 & 13.95 & 26.20          & 74.32 & 80.83 & 69.19 & 31.67 & 64.00 \\
\midrule
\multirow{4}{*}{qwen-plus} 
 & RAG   & 39.40 & 74.44 & 33.32 & 37.14 & \underline{46.18} & 77.18 & 92.12 & 72.30 & 59.34 & \textbf{75.16} \\
 & A-Mem & 43.78 & 78.97 & 38.25 & 36.67 & \textbf{49.54}    & 75.34 & 92.12 & 67.60 & 63.31 & \underline{74.51} \\
 & Mem0  & 16.59 & 36.97 & 11.96 & 15.18 & 20.20             & 72.91 & 72.79 & 59.85 & 48.36 & 63.46 \\
 & MemOS & 38.86 & 70.68 & 26.59 & 23.40 & 39.88             & 74.54 & 86.36 & 67.95 & 52.95 & 70.45 \\
\bottomrule
\end{tabular}
}
\end{table*}

\section{Experiments}
\subsection{Experimental Setup}
\label{experiment_setup}
To assess the performance of current mainstream memory systems in the context of continuous dialogue lifelogs and to derive insights, we select four representative memory systems for evaluation on \textsc{LifeDialBench}: 
(1) \textbf{RAG}~\citep{rag}: A straightforward retrieval-augmented generation (RAG) baseline that directly stores and retrieves text chunks;
(2) \textbf{A-Mem}~\citep{amem}: An enhanced variant of RAG that augments the retrieval process with additional semantic signals — such as context, tags, keywords, and links — to improve the representation and utilization of stored text;
(3) \textbf{MemOS}~\citep{memos}: A memory system that does not retain raw input, but instead abstracts each input segment into a single structured representation—including summaries, titles, and semantic tags;
and (4) \textbf{Mem0}~\citep{mem0}: A memory paradigm that similarly discards raw input, but extracts and stores multiple concise factual statements per input segment. More information about the implementation details of these methods can be found in \Cref{app:memory_system}.

Experiments on EgoMem and LifeMem are conducted in an online setting using \texttt{GPT-4o-mini}
\footnote{Corresponds to \href{https://chatgpt.com/zh-Hans-CN/overview?openaicom_referred=true}{\texttt{GPT-4o-mini}}}
and \texttt{qwen-plus}
\footnote{Corresponds to \href{https://bailian.console.aliyun.com/cn-beijing?spm=5176.29619931.nav-v2-dropdown-menu-0.d_main_2_0.74cd10d7sBHNbm&tab=model&scm=20140722.M_10944401._.V_1\#/model-market/detail/qwen-plus-2025-12-01}{\texttt{qwen-plus-2025-12-01}}}

To further investigate the necessity of online evaluation, we also include \texttt{Qwen3-8B} in additional experiments. For embedding needs, we employ \texttt{Qwen3-Embedding-8B}.

We adopt two question-answering formats: multiple-choice and open-ended. For multiple-choice questions, correctness is determined by exact matching. For open-ended questions, the reference answer is constructed from the correct option of the corresponding multiple-choice question. We then use \texttt{Qwen3-32B} as an evaluator to judge whether the generated response is semantically equivalent to the reference.

\subsection{Main Results}
\label{sec:main_results}
To assess the performance of existing memory systems in the context of continuous dialogue lifelogs, we evaluate the four representative systems mentioned in \S\ref{experiment_setup} on \textsc{LifeDialBench}. \Cref{tab:main_results} presents the full results. In the following, we analyze three aspects: (1) the importance of storing raw text, (2) the impact of compression level on performance, and (3) the difference in results between different question settings.

\paragraph{Importance of Raw Text}
Contrary to the prevailing trend favoring complex, structured memory architectures, our results reveal a counter-intuitive finding in the lifelog domain: simple raw-text preservation (RAG, A-Mem) significantly outperforms sophisticated summarization-based paradigms (Mem0, MemOS).

Specifically, RAG and A-Mem consistently achieve the highest performance across both subsets and backbone models (\texttt{GPT-4o-mini} and \texttt{Qwen-Plus}). While A-Mem represents an ``enhanced'' paradigm that augments raw text with auxiliary metadata (brief summaries and associations), it does not yield statistically significant gains over the vanilla RAG baseline. This suggests that for continuous dialogue streams, the fidelity of the original context is the dominant factor for retrieval success, rendering added structural complexity largely redundant.

\paragraph{Impact of Compression Level}
MemOS and Mem0 represent two memory approaches that compress raw text to varying extents. While MemOS applies lightweight summarization to the original dialogue, Mem0 performs fact-level compression, resulting in compression ratios of 62\% and 35\% in terms of token count, respectively. Our results show that MemOS, despite underperforming compared to the uncompressed baselines RAG and A-Mem, still substantially outperforms Mem0. This indicates that compressing raw text not only degrades performance, but also that the degree of compression correlates with the extent of performance loss.

\paragraph {Impact of Question Type} By comparing system performance on four question categories, we find that Event Detail Retrieval (QT2) is the relatively easiest task, followed by Event Content Recall (QT1), while Multi-hop Event Reasoning (QT3) and Temporal Grounding (QT4) prove to be the most challenging. The relative ease of QT2 is intuitive, as it merely requires retrieving specific details from the original text. QT1, by contrast, demands that the agent infer the occurred event from dialogue content, making it comparatively more difficult. The low performance on QT3 and QT4 suggests that multi-hop reasoning and temporal grounding remain the most formidable challenges in the context of continuous dialogue lifelogs.

\paragraph {Impact of QA Format} Comparing the Open-Ended and Multiple-Choice settings reveals that questions are considerably more difficult in the Open-Ended format. Notably, while QT3 outperforms QT4 in the Multiple-Choice setting, this trend reverses in the Open-Ended setting. This suggests that without the guidance of candidate options, models struggle to aggregate and reason over disjoint evidence from the continuous stream, indicating that independent reasoning imposes higher demands on contextual fidelity.

\section{Analysis of Online Evaluation}

To quantify the "future context contamination" defined in \cref{sec:online_eval}, we analyzed the correlation between the retrieval of future memories and answer correctness using RAG and A-Mem on LifeMem. We computed the AUROC scores for this relationship, obtaining \textbf{0.64} and \textbf{0.68}, respectively. These non-random values ($>0.5$) demonstrate a dependency: the presence of future information materially distorts model performance. Whether this leakage acts as a "cheat" or "noise", it creates an uncontrolled confound that renders offline metrics unreliable for assessing real-world utility.

Mem0 exemplifies a memory system with overwrite-based updates. To demonstrate the impact of "irreversible memory modification". We identified a set of questions that Mem0 answers correctly under the Online setting but fails in the offline setting on LifeMem. Even after restricting the offline retriever to memories created before the query timestamp and increasing the retrieval budget (top-$k$) from 20 to 100, the accuracy on this subset remains only \textbf{34.91\%}—far below 100\%. This strongly suggests that the required information was likely overwritten during the offline memory construction process.

\begin{figure}[t]
    \centering
    \includegraphics[width=1\linewidth]{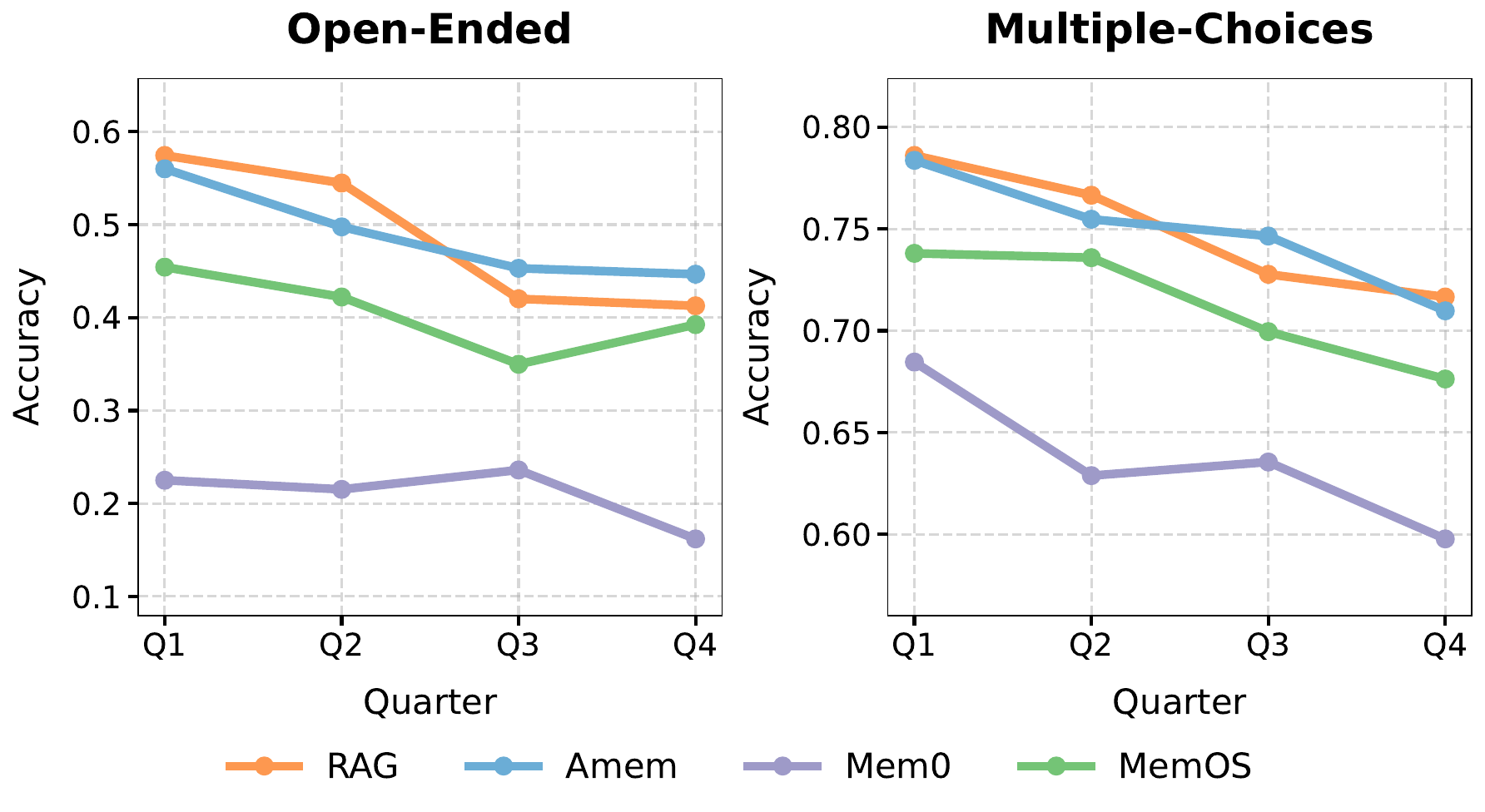}
    \caption{Comparison of accuracy decay rates across memory systems. RAG exhibits the steepest decline, whereas abstraction-based methods (A-Mem, MemOS) maintain better stability over long-term timelines despite lower absolute accuracy.}
    \label{fig:acc_trend}
    \vspace{-12pt}
\end{figure}

\paragraph{Accuracy Decay over Time.}
This is a characteristic feature of online evaluation. \Cref{fig:acc_trend} illustrates the quarterly accuracy of \texttt{Qwen-Plus} on LifeMem across different quarters. All systems exhibit a declining trend, as the memory pool continuously expands under the online setting, increasing the difficulty of retrieval. This phenomenon aligns with real-world scenarios.

\section{Conclusion}

In this paper, we bridge the gap between isolated dialogue sessions and continuous real-world existence by introducing \textbf{\textsc{LifeDialBench}}, a comprehensive benchmark comprising the real-world EgoMem and the synthesized year-long LifeMem. Crucially, we identify that traditional offline evaluation fundamentally violates the temporal causality of lifelogs, prompting the proposal of a rigorous Online Evaluation protocol. 

Our extensive experiments yield a counter-intuitive yet critical finding: despite the growing complexity of agentic memory architectures (e.g., graph-based or summary-based systems), simple raw-text retrieval (RAG) consistently outperforms structured memory paradigms. Our analysis reveals that current abstraction mechanisms introduce "lossy compression," stripping away vital contextual details required for precise grounding in continuous streams. Furthermore, we demonstrate that offline metrics suffer from severe future context contamination, rendering them unreliable. By establishing these insights, \textsc{LifeDialBench} serves not only as a testbed but as a call to action for the community to rethink memory design—shifting focus from aggressive abstraction back to high-fidelity context preservation.

\section*{Ethics Statement}

Wearable devices with always-on microphones are changing the balance between personal memory tools and unwanted monitoring. While continuous lifelogging offers new possibilities for personalized AI, it brings serious ethical questions around privacy, consent, and data protection.

\paragraph{Consent and Bystander Privacy.} 
A primary concern of always-on lifelogging is the inadvertent capture of non-consenting third parties. Unlike voluntary human-AI interactions, continuous logs inherently record speech from colleagues, family, or strangers, raising significant legal and ethical issues under frameworks such as the GDPR and various wiretapping laws. \textsc{LifeDialBench} is constructed to sidestep these concerns: \textbf{EgoMem} draws from the publicly released and ethically vetted EgoLife dataset, while \textbf{LifeMem} employs human-in-the-loop simulation within a virtual community. For real-world deployment, we advocate for robust consent mechanisms, including visible recording indicators and granular opt-out affordances for bystanders.

\paragraph{Data Security and Sensitive Information.} 
The temporal depth and granularity inherent in longitudinal lifelogging records, which encompass sensitive information from healthcare, personal finance, and social relationships, significantly amplifies the potential for adversarial behavioral profiling. To prevent unauthorized access to a user's complete social graph or psychological state, a privacy-by-design architecture rooted in the edge-intelligence paradigm offers an effective solution. Raw audio capture can be processed entirely on-device within secure enclaves, with transcripts and vector embeddings confined to encrypted, sandboxed local storage. Furthermore, enforcing user-configurable data retention policies (e.g., rolling deletion of raw data within 24 hours) minimizes the persistent attack surface and renders centralized large-scale data exfiltration technically infeasible.

\section*{Limitations}
\label{sec:limitation}

We acknowledge three primary limitations in \textsc{LifeDialBench}. 

First, the year-long LifeMem relies on simulation and lacks the stochasticity of raw ASR noise. However, we argue this abstraction is justifiable as it aligns with modern pipelines where ASR outputs typically undergo upstream refinement. Benchmarking on high-fidelity text allows us to strictly isolate memory reasoning bottlenecks from perceptual errors, establishing a critical upper-bound baseline.

Second, our benchmark is currently restricted to the textual modality. The exclusion of visual cues limits the evaluation of tasks requiring visual grounding (e.g., locating objects), which we leave for future multimodal extensions.

Finally, the computational intensity of Online Evaluation limited our experiments to representative backbone models. This precluded an exhaustive exploration of emerging ultra-long-context models in no-retrieval settings, which remains an open avenue for future research.

\bibliography{custom}

@misc{mudarisov2025limitationsnormalizationattentionmechanism,
      title={Limitations of Normalization in Attention Mechanism}, 
      author={Timur Mudarisov and Mikhail Burtsev and Tatiana Petrova and Radu State},
      year={2025},
      eprint={2508.17821},
      archivePrefix={arXiv},
      primaryClass={cs.LG},
      url={https://arxiv.org/abs/2508.17821}, 
}

@InProceedings{EgoLife_Yang_2025_CVPR,
    author    = {Yang, Jingkang and Liu, Shuai and Guo, Hongming and Dong, Yuhao and Zhang, Xiamengwei and Zhang, Sicheng and Wang, Pengyun and Zhou, Zitang and Xie, Binzhu and Wang, Ziyue and Ouyang, Bei and Lin, Zhengyu and Cominelli, Marco and Cai, Zhongang and Li, Bo and Zhang, Yuanhan and Zhang, Peiyuan and Hong, Fangzhou and Widmer, Joerg and Gringoli, Francesco and Yang, Lei and Liu, Ziwei},
    title     = {EgoLife: Towards Egocentric Life Assistant},
    booktitle = {Proceedings of the IEEE/CVF Conference on Computer Vision and Pattern Recognition (CVPR)},
    month     = {June},
    year      = {2025},
    pages     = {28885-28900}
}

@misc{openai2022chatgpt,
  author    = {OpenAI},
  title     = {ChatGPT},
  year      = {2022},
  url       = {https://chat.openai.com/chat},
 note      = {Accessed: September 15, 2024},
}

@inproceedings{memorybank,
  author       = {Wanjun Zhong and
                  Lianghong Guo and
                  Qiqi Gao and
                  He Ye and
                  Yanlin Wang},
  editor       = {Michael J. Wooldridge and
                  Jennifer G. Dy and
                  Sriraam Natarajan},
  title        = {MemoryBank: Enhancing Large Language Models with Long-Term Memory},
  booktitle    = {Thirty-Eighth {AAAI} Conference on Artificial Intelligence, {AAAI}
                  2024, Thirty-Sixth Conference on Innovative Applications of Artificial
                  Intelligence, {IAAI} 2024, Fourteenth Symposium on Educational Advances
                  in Artificial Intelligence, {EAAI} 2014, February 20-27, 2024, Vancouver,
                  Canada},
  pages        = {19724--19731},
  publisher    = {{AAAI} Press},
  year         = {2024},
  url          = {https://doi.org/10.1609/aaai.v38i17.29946},
  doi          = {10.1609/AAAI.V38I17.29946},
  bibsource    = {dblp computer science bibliography, https://dblp.org}
}

@misc{text-grad,
      title={TextGrad: Automatic "Differentiation" via Text}, 
      author={Mert Yuksekgonul and Federico Bianchi and Joseph Boen and Sheng Liu and Zhi Huang and Carlos Guestrin and James Zou},
      year={2024},
      eprint={2406.07496},
      archivePrefix={arXiv},
      primaryClass={cs.CL},
      url={https://arxiv.org/abs/2406.07496}, 
}

@misc{rag,
      title={Retrieval-Augmented Generation for Knowledge-Intensive NLP Tasks}, 
      author={Patrick Lewis and Ethan Perez and Aleksandra Piktus and Fabio Petroni and Vladimir Karpukhin and Naman Goyal and Heinrich Küttler and Mike Lewis and Wen-tau Yih and Tim Rocktäschel and Sebastian Riedel and Douwe Kiela},
      year={2021},
      eprint={2005.11401},
      archivePrefix={arXiv},
      primaryClass={cs.CL},
      url={https://arxiv.org/abs/2005.11401}, 
}

@misc{membench_gaoling,
      title={MemBench: Towards More Comprehensive Evaluation on the Memory of LLM-based Agents}, 
      author={Haoran Tan and Zeyu Zhang and Chen Ma and Xu Chen and Quanyu Dai and Zhenhua Dong},
      year={2025},
      eprint={2506.21605},
      archivePrefix={arXiv},
      primaryClass={cs.CL},
      url={https://arxiv.org/abs/2506.21605}, 
}

@inproceedings{llm-as-a-judge,
 author = {Zheng, Lianmin and Chiang, Wei-Lin and Sheng, Ying and Zhuang, Siyuan and Wu, Zhanghao and Zhuang, Yonghao and Lin, Zi and Li, Zhuohan and Li, Dacheng and Xing, Eric and Zhang, Hao and Gonzalez, Joseph E and Stoica, Ion},
 booktitle = {Advances in Neural Information Processing Systems},
 editor = {A. Oh and T. Naumann and A. Globerson and K. Saenko and M. Hardt and S. Levine},
 pages = {46595--46623},
 publisher = {Curran Associates, Inc.},
 title = {Judging LLM-as-a-Judge with MT-Bench and Chatbot Arena},
 url = {https://proceedings.neurips.cc/paper_files/paper/2023/file/91f18a1287b398d378ef22505bf41832-Paper-Datasets_and_Benchmarks.pdf},
 volume = {36},
 year = {2023}
}

@inproceedings{locomo,
    title = "Evaluating Very Long-Term Conversational Memory of {LLM} Agents",
    author = "Maharana, Adyasha  and
      Lee, Dong-Ho  and
      Tulyakov, Sergey  and
      Bansal, Mohit  and
      Barbieri, Francesco  and
      Fang, Yuwei",
    editor = "Ku, Lun-Wei  and
      Martins, Andre  and
      Srikumar, Vivek",
    booktitle = "Proceedings of the 62nd Annual Meeting of the Association for Computational Linguistics (Volume 1: Long Papers)",
    month = aug,
    year = "2024",
    address = "Bangkok, Thailand",
    publisher = "Association for Computational Linguistics",
    url = "https://aclanthology.org/2024.acl-long.747/",
    doi = "10.18653/v1/2024.acl-long.747",
    pages = "13851--13870"
}

@article{amem,
  title={A-mem: Agentic memory for llm agents},
  author={Xu, Wujiang and Liang, Zujie and Mei, Kai and Gao, Hang and Tan, Juntao and Zhang, Yongfeng},
  journal={arXiv preprint arXiv:2502.12110},
  year={2025}
}

@misc{mem0,
      title={Mem0: Building Production-Ready AI Agents with Scalable Long-Term Memory}, 
      author={Prateek Chhikara and Dev Khant and Saket Aryan and Taranjeet Singh and Deshraj Yadav},
      year={2025},
      eprint={2504.19413},
      archivePrefix={arXiv},
      primaryClass={cs.CL},
      url={https://arxiv.org/abs/2504.19413}, 
}

@inproceedings{hipporag,
author = {Guti\'{e}rrez, Bernal Jim\'{e}nez and Shu, Yiheng and Gu, Yu and Yasunaga, Michihiro and Su, Yu},
title = {HippoRAG: neurobiologically inspired long-term memory for large language models},
year = {2025},
isbn = {9798331314385},
publisher = {Curran Associates Inc.},
address = {Red Hook, NY, USA},
booktitle = {Proceedings of the 38th International Conference on Neural Information Processing Systems},
articleno = {1902},
numpages = {38},
location = {Vancouver, BC, Canada},
series = {NIPS '24}
}

@misc{personamem,
      title={Know Me, Respond to Me: Benchmarking LLMs for Dynamic User Profiling and Personalized Responses at Scale}, 
      author={Bowen Jiang and Zhuoqun Hao and Young-Min Cho and Bryan Li and Yuan Yuan and Sihao Chen and Lyle Ungar and Camillo J. Taylor and Dan Roth},
      year={2025},
      eprint={2504.14225},
      archivePrefix={arXiv},
      primaryClass={cs.CL},
      url={https://arxiv.org/abs/2504.14225}, 
}

@article{memgpt,
  publtype={informal},
  author={Charles Packer and Vivian Fang and Shishir G. Patil and Kevin Lin and Sarah Wooders and Joseph E. Gonzalez},
  title={MemGPT: Towards LLMs as Operating Systems},
  year={2023},
  cdate={1672531200000},
  journal={CoRR},
  volume={abs/2310.08560},
  url={https://doi.org/10.48550/arXiv.2310.08560}
}

@article{memos,
  title={Memos: A memory os for ai system},
  author={Li, Zhiyu and Song, Shichao and Xi, Chenyang and Wang, Hanyu and Tang, Chen and Niu, Simin and Chen, Ding and Yang, Jiawei and Li, Chunyu and Yu, Qingchen and others},
  journal={arXiv preprint arXiv:2507.03724},
  year={2025}
}

@inproceedings{longmemeval,
  title={LongMemEval: Benchmarking Chat Assistants on Long-Term Interactive Memory},
  author={Wu, Di and Wang, Hongwei and Yu, Wenhao and Zhang, Yuwei and Chang, Kai-Wei and Yu, Dong},
  booktitle={The Thirteenth International Conference on Learning Representations},
  year={2024}
}

@article{zep,
  title={Zep: a temporal knowledge graph architecture for agent memory},
  author={Rasmussen, Preston and Paliychuk, Pavlo and Beauvais, Travis and Ryan, Jack and Chalef, Daniel},
  journal={arXiv preprint arXiv:2501.13956},
  year={2025}
}

@misc{scm,
      title={SCM: Enhancing Large Language Model with Self-Controlled Memory Framework}, 
      author={Bing Wang and Xinnian Liang and Jian Yang and Hui Huang and Shuangzhi Wu and Peihao Wu and Lu Lu and Zejun Ma and Zhoujun Li},
      year={2025},
      eprint={2304.13343},
      archivePrefix={arXiv},
      primaryClass={cs.CL},
      url={https://arxiv.org/abs/2304.13343}, 
}

@article{egor1,
  title={Ego-R1: Chain-of-Tool-Thought for Ultra-Long Egocentric Video Reasoning},
  author={Tian, Shulin and Wang, Ruiqi and Guo, Hongming and Wu, Penghao and Dong, Yuhao and Wang, Xiuying and Yang, Jingkang and Zhang, Hao and Zhu, Hongyuan and Liu, Ziwei},
  journal={arXiv preprint arXiv:2506.13654},
  year={2025}
}

@misc{qwen3,
      title={Qwen3 Technical Report}, 
      author={An Yang and Anfeng Li and Baosong Yang and Beichen Zhang and Binyuan Hui and Bo Zheng and Bowen Yu and Others},
      year={2025},
      eprint={2505.09388},
      archivePrefix={arXiv},
      primaryClass={cs.CL},
      url={https://arxiv.org/abs/2505.09388}, 
}

@misc{gpt4,
      title={GPT-4 Technical Report}, 
      author={OpenAI and Josh Achiam and Steven Adler and Sandhini Agarwal and Lama Ahmad and Ilge Akkaya and Florencia Leoni Aleman and Diogo Almeida and Janko Altenschmidt and Sam Altman and Others},
      year={2024},
      eprint={2303.08774},
      archivePrefix={arXiv},
      primaryClass={cs.CL},
      url={https://arxiv.org/abs/2303.08774}, 
}

@misc{naih,
	author = {NAIH},
	title = {{G}it{H}ub - gkamradt/{L}{L}{M}{T}est\_{N}eedle{I}n{A}{H}aystack: {D}oing simple retrieval from {L}{L}{M} models at various context lengths to measure accuracy --- github.com},
	howpublished = {\url{https://github.com/gkamradt/LLMTest_NeedleInAHaystack}},
	year = {2025},
	note = {[Accessed 18-09-2025]},
}

@misc{longlora,
      title={LongLoRA: Efficient Fine-tuning of Long-Context Large Language Models}, 
      author={Yukang Chen and Shengju Qian and Haotian Tang and Xin Lai and Zhijian Liu and Song Han and Jiaya Jia},
      year={2024},
      eprint={2309.12307},
      archivePrefix={arXiv},
      primaryClass={cs.CL},
      url={https://arxiv.org/abs/2309.12307}, 
}

@misc{llama3,
      title={The Llama 3 Herd of Models}, 
      author={Aaron Grattafiori and Abhimanyu Dubey and Abhinav Jauhri and Abhinav Pandey and Abhishek Kadian and Ahmad Al-Dahle and Aiesha Letman and Akhil Mathur and Alan Schelten and Others},
      year={2024},
      eprint={2407.21783},
      archivePrefix={arXiv},
      primaryClass={cs.AI},
      url={https://arxiv.org/abs/2407.21783}, 
}

@inproceedings{
ruler,
title={{RULER}: What{\textquoteright}s the Real Context Size of Your Long-Context Language Models?},
author={Cheng-Ping Hsieh and Simeng Sun and Samuel Kriman and Shantanu Acharya and Dima Rekesh and Fei Jia and Boris Ginsburg},
booktitle={First Conference on Language Modeling},
year={2024},
url={https://openreview.net/forum?id=kIoBbc76Sy}
}

@inproceedings{loogle,
    title = "{L}oo{GLE}: Can Long-Context Language Models Understand Long Contexts?",
    author = "Li, Jiaqi  and
      Wang, Mengmeng  and
      Zheng, Zilong  and
      Zhang, Muhan",
    editor = "Ku, Lun-Wei  and
      Martins, Andre  and
      Srikumar, Vivek",
    booktitle = "Proceedings of the 62nd Annual Meeting of the Association for Computational Linguistics (Volume 1: Long Papers)",
    month = aug,
    year = "2024",
    address = "Bangkok, Thailand",
    publisher = "Association for Computational Linguistics",
    url = "https://aclanthology.org/2024.acl-long.859/",
    doi = "10.18653/v1/2024.acl-long.859",
    pages = "16304--16333"
}

@article{lost-in-the-middle,
    title = "Lost in the Middle: How Language Models Use Long Contexts",
    author = "Liu, Nelson F.  and
      Lin, Kevin  and
      Hewitt, John  and
      Paranjape, Ashwin  and
      Bevilacqua, Michele  and
      Petroni, Fabio  and
      Liang, Percy",
    journal = "Transactions of the Association for Computational Linguistics",
    volume = "12",
    year = "2024",
    address = "Cambridge, MA",
    publisher = "MIT Press",
    url = "https://aclanthology.org/2024.tacl-1.9/",
    doi = "10.1162/tacl_a_00638",
    pages = "157--173"
}

@article{toolformer,
  title={Toolformer: Language models can teach themselves to use tools},
  author={Schick, Timo and Dwivedi-Yu, Jane and Dess{\`\i}, Roberto and Raileanu, Roberta and Lomeli, Maria and Hambro, Eric and Zettlemoyer, Luke and Cancedda, Nicola and Scialom, Thomas},
  journal={Advances in Neural Information Processing Systems},
  volume={36},
  pages={68539--68551},
  year={2023}
}

@article{gpt4tools,
  title={Gpt4tools: Teaching large language model to use tools via self-instruction},
  author={Yang, Rui and Song, Lin and Li, Yanwei and Zhao, Sijie and Ge, Yixiao and Li, Xiu and Shan, Ying},
  journal={Advances in Neural Information Processing Systems},
  volume={36},
  pages={71995--72007},
  year={2023}
}

@article{cheng2024videgothink,
  title={Videgothink: Assessing egocentric video understanding capabilities for embodied ai},
  author={Cheng, Sijie and Fang, Kechen and Yu, Yangyang and Zhou, Sicheng and Li, Bohao and Tian, Ye and Li, Tingguang and Han, Lei and Liu, Yang},
  journal={arXiv preprint arXiv:2410.11623},
  year={2024}
}

\onecolumn
\newpage
\appendix
\section{LLM Usage}
In the preparation of this paper, large language models (LLMs) were used solely as auxiliary tools. Specifically, we employed LLMs for grammar correction and text polishing, as well as to support dataset generation and assist in the manual review of data quality.

\section{Details of human-in-the-loop review}
\label{app:human-in-the-loop}
\paragraph{Overall Procedure and LLM-Assisted Inspection.} The overall procedure begins with \textit{monthly-level summary}, where annotators perform comprehensive reading, inspection, and revision. This is followed by \textit{weekly-level summary}, which involves several checks: \textbf{(2.1) Consistency between parent and child summaries} verifies that weekly content does not contradict monthly content (e.g., ensuring events are not mistakenly placed before a meeting); \textbf{(2.2) Factual correctness} checks for obvious factual errors (e.g., accurately reflecting the initials on a ring); \textbf{(2.3) Repetition checking} uses an LLM to extract event descriptions, retrieves the five most similar events via similarity search, and inspects them to prevent unreasonable repetition (e.g., the protagonist reading the same book chapter and having identical reflections in different months); and \textbf{(2.4) Random sampling}, where 20 revised summaries are randomly selected for additional verification. The \textit{day-level and event-level summaries} follow the same checking procedure as the weekly-level summaries. Given the substantial volume of content at the weekly, day, and event levels, we employ an LLM (specifically, \texttt{Qwen3-235B-Instruct}) to conduct a first-pass review for detecting potentially problematic segments, after which human annotators perform full manual inspection and correction.

\paragraph{Issue Detection and Final Quality.} The proportions of issues detected by the LLM during the initial review are as follows: for parent-child consistency, weekly (11\%), day (14\%), and event (16\%); for factual correctness, weekly (13\%), day (17\%), and event (19\%); and for repetition checking, weekly (7\%), day (10\%), and event (12\%). Following revisions based on these checks and subsequent human review, the final quality is confirmed through random sampling, which yields a pass rate of 100\% for the weekly, day, and event-level summaries. All review work is conducted by several data annotators within our team.

\section{Annotator Details}
\label{app:annotator}

All human annotators are professional data annotation employees of RayNeo AI, recruited internally from the company's dedicated annotation team. Compensation was provided in accordance with the company's standard salary structure, commensurate with their professional roles.

\section{Details of EgoMem Construction}\label{app:egomem_details}

In this section, we provide a comprehensive breakdown of the methodological decisions behind the construction of the EgoMem benchmark, specifically addressing the limitations of direct audio transcription, the rationale for our sliding-window generation strategy, and the human-in-the-loop review process designed to ensure conversational naturalness.

\subsection{Limitations of Direct ASR Transcription}\label{app:egomem_asr}
Our initial objective for EgoMem was to utilize Automatic Speech Recognition (ASR) to directly transcribe the raw audio captured by the Meta Aria smart glasses in the EgoLife dataset~\citep{EgoLife_Yang_2025_CVPR}. However, empirical investigations revealed several critical bottlenecks that rendered this approach impractical for constructing a robust long-term memory benchmark:
\begin{itemize}
    \item \textbf{Data Sparsity:} Although EgoLife contains over 300 hours of recording (collected from six participants over seven days), the actual density of meaningful, continuous conversations is remarkably sparse. Direct transcription resulted in vast segments of silence or fragmented ambient noise, which is insufficient for evaluating complex memory retrieval and reasoning over long contexts.
    \item \textbf{ASR Degradation in the Wild:} Real-world egocentric audio is highly unstructured. It is plagued by severe background noise (e.g., wind, traffic, appliances), overlapping speech from multiple speakers, and varying distances from the microphone. State-of-the-art ASR systems struggled to produce coherent transcripts under these conditions, introducing severe perceptual errors that would confound the evaluation of an agent's memory reasoning capabilities (i.e., failing to answer a query due to ASR garbling rather than memory retrieval failure).
    \item \textbf{Incompatibility of Existing Q\&A Pairs:} While the original EgoLife dataset provides its own human-annotated Q\&A pairs, these queries are inherently \textit{video-grounded} (e.g., asking about the visual state of an object or spatial relations). Because our benchmark specifically targets text-centric conversational memory architectures, these visual queries are not directly applicable. 
\end{itemize}
Consequently, we pivoted to a hybrid proxy approach: using the high-fidelity, VLM-generated 10-minute visual summaries from Ego-R1~\citep{egor1} as factual anchors, and prompting an LLM to synthesize the corresponding conversational interactions. This approach isolates memory reasoning bottlenecks from perceptual (ASR/Vision) errors while preserving the true physical event timeline.

\subsection{Justification for Granularity and Context Window}\label{app:egomem_granularity}
A core challenge in generating summary-grounded simulated dialogues is selecting the appropriate temporal granularity. Ego-R1 provides summaries at multiple scales (30-second, 10-minute, 1-hour, 1-day, 1-week). 
\begin{itemize}
    \item \textbf{Granularity Selection:} We deliberately selected the 10-minute summaries. Finer-grained 30-second summaries are too fragmented, leading to disjointed micro-interactions that lack conversational flow. Conversely, coarser summaries (1-hour or 1-day) abstract away critical episodic details, resulting in superficial dialogues. We observed that asking an LLM to generate continuous dialogue directly from an hour-long summary often leads to severe information loss, a phenomenon consistent with the limited effective attention span of current LLMs when processing dense event descriptions~\citep{mudarisov2025limitationsnormalizationattentionmechanism}.
    \item \textbf{Sliding-Window Strategy:} To bridge discrete 10-minute segments into a continuous daily lifelog without losing temporal coherence, we designed a sliding-window generation strategy. When generating the dialogue for segment $T$, the LLM is conditioned on the textual summaries of segments $T-6$ to $T$ (representing a 60-minute historical context). This 60-minute window was empirically found to be the optimal sweet spot: it provides sufficient context to maintain speaker consistency, pronoun resolution, and ongoing topic flow, while preventing the LLM from entering hallucination loops or repeating distant historical interactions.
\end{itemize}

\subsection{Iterative Review and Conversational Grounding}\label{app:egomem_review}
To prevent the generated dialogues from reading like overly formalized written prose---a common artifact of standard LLM generation---we implemented a rigorous human-LLM collaborative review pipeline inspired by \textit{text-grad}~\citep{text-grad}. The explicit goal of this pipeline was to optimize for \textbf{Conversational Naturalness}.

\begin{enumerate}
    \item \textbf{LLM-Assisted Flagging:} We deployed a critique LLM instructed to evaluate the generated dialogues. Instead of optimizing for grammatical perfection, the critique model flagged segments that sounded ``too formal,'' ``structurally rigid,'' or ``unnatural for close acquaintances.''
    \item \textbf{Human Revision for Naturalness:} Human annotators reviewed the flagged segments and revised them to mimic real-world continuous recordings. Key adjustments included:
    \begin{itemize}
        \item \textit{Casual Phrasing and Tone:} Replacing highly structured, essay-like sentences with relaxed, colloquial expressions typical of daily roommate or family interactions.
        \item \textit{Implicit Contexts:} Ensuring that characters do not overly explain situations that close acquaintances would implicitly understand (e.g., referring to ``the project'' rather than ``the software engineering project we discussed yesterday''). This forces the evaluated memory systems to rely on long-term context resolution rather than immediate explicit clues.
        \item \textit{Dynamic Flow:} Smoothing out the transitions between the sliding windows to ensure topic shifts felt spontaneous rather than programmatic.
    \end{itemize}
    \item \textbf{Final Factual Alignment:} The revised, naturalized dialogues were checked one final time against the Ego-R1 summaries to guarantee that no physical events or critical factual details were altered during the conversational grounding process.
\end{enumerate}

\section{Sensitivity Analysis}
\subsection{Impact of Backbone Capability} 
\begin{figure*}[h]
    \vspace{-10pt}
    \centering
    \includegraphics[width=\linewidth]{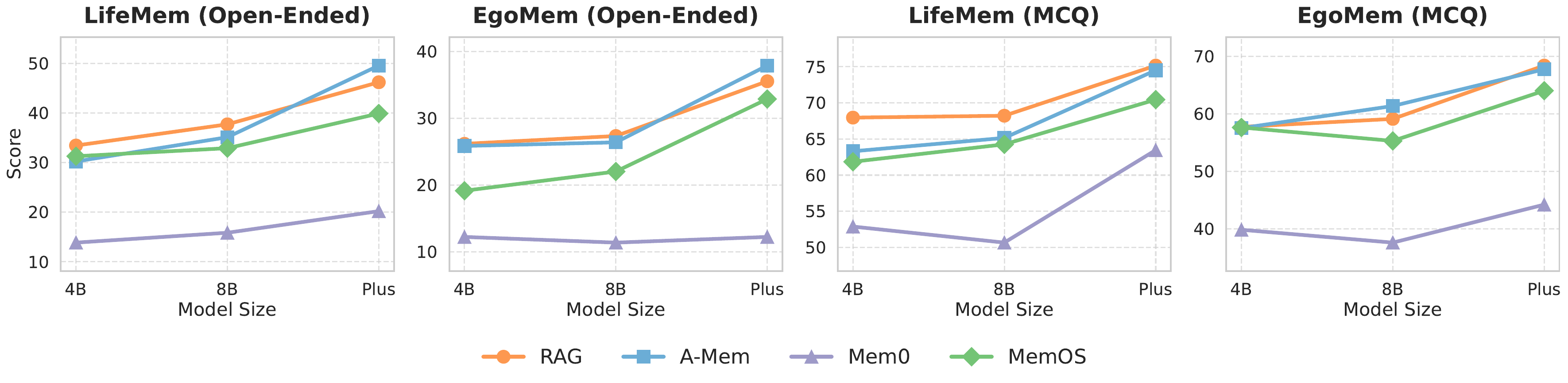}
    \vspace{-20pt}
    \caption{Comparison on the performances of memory systems using different backbone LLMs.}
    \label{fig:model_size}
    \vspace{-10pt}
\end{figure*}

We investigate the influence of the underlying model's capacity on memory performance by conducting experiments with \texttt{Qwen3-4B}, \texttt{Qwen3-8B}, and \texttt{Qwen-Plus}. As illustrated in Figure~\ref{fig:model_size}, the results demonstrate that the overall effectiveness of the memory system is positively correlated with the inherent capabilities of the base model. Notably, the transition from the 4B to the 8B variant yields only marginal performance gains and occasionally results in slight performance degradation. In contrast, upgrading to the \texttt{qwen-plus} model leads to a substantial improvement across evaluated metrics. These findings suggest that the deployment of high-capacity models is essential for maximizing the utility of agentic memory systems in complex tasks.

\subsection{Sensitivity to Retrieval Top-K.}
To investigate the influence of retrieval breadth on system efficacy, we conduct an ablation study on LifeMem leveraging \texttt{Qwen3-8B} under varying top-$k$ configurations ($k \in \{10, 20, 40\}$). As illustrated in Table \ref{tab:topk}, performance generally scales positively with the retrieval depth $k$. Although A-Mem exhibits a minor performance regression in the Open-Ended task when $k$ increases from 20 to 40, the results across both models and tasks consistently demonstrate that larger $k$ values yield superior outcomes. This prevailing trend underscores that retrieval recall remains a primary bottleneck for memory-augmented agents. Furthermore, the findings suggest substantial potential for enhancing downstream performance by optimizing the retrieval modules of contemporary memory systems to more effectively manage broader context.

\begin{table}[h]
\centering
\setlength{\tabcolsep}{6pt} 
\renewcommand{\arraystretch}{1} 
\caption{Performance sensitivity analysis across varying top-$k$ retrieval configurations on LifeMem.} 
\label{tab:topk}
\begin{tabular}{cc|ccc} 
\toprule
\multirow{2}{*}{Method} & \multirow{2}{*}{Task} & \multicolumn{3}{c}{Top-$k$} \\ 
\cmidrule(l){3-5} 
 &  & 10 & 20 & 40 \\ 
\midrule
\multirow{2}{*}{A-Mem} & Open-Ended & 29.06 & 35.10 & 34.89 \\
 & MCQ & 61.90 & 65.15 & 68.46 \\ 
\midrule 
\multirow{2}{*}{MemOS} & Open-Ended & 27.07 & 32.89 & 33.43 \\
 & MCQ & 59.10 & 64.26 & 67.79 \\ 
\bottomrule
\end{tabular}
\end{table}

\section{Benchmark Data Samples}
\label{app:data_example}
\subsection{Examples of dataset}
In this section, we provide illustrative snippets from the LifeMem Dataset. The EgoMem Dataset adopts the same formatting and structural schema.

\renewcommand{\arraystretch}{1.3}

\begin{longtable}{p{2cm} >{\bfseries}p{1cm} p{11cm}}
    \hiderowcolors
    \caption{Jeremy and Jane at Home Organizing Old Items (2024-01-01)} \label{tab:dialogue_sample} \\
    \toprule
    \rowcolor{white}
    \textbf{Time} & \textbf{Speaker} & \textbf{Utterance} \\
    \midrule
    \endfirsthead
    
    \multicolumn{3}{c}%
    {{\bfseries \tablename\ \thetable{} -- continued from previous page}} \\
    \toprule
    \textbf{Time} & \textbf{Speaker} & \textbf{Utterance} \\
    \midrule
    \endhead
    
    \bottomrule
    \multicolumn{3}{r}{{Continued on next page...}} \\
    \endfoot
    
    \bottomrule
    \endlastfoot
    
    \ttfamily [08:10:15] & Jane & All done eating. I'll go clear the bowls first, and then shall we tidy up the cabinet in the living room? \\
    \ttfamily [08:10:22] & Jeremy & Okay, I'll help you clear up. No point letting them pile up. \\
    \ttfamily [08:10:30] & Jane & Yeah, and give the tablecloth a good shake while you're at it, there are some breadcrumbs. \\
    \ttfamily [08:10:38] & Jeremy & Alright, you go change into some clothes you don't mind getting dirty. I'll be over as soon as I finish here. \\
    \ttfamily [08:11:05] & Jane & Hey, the bottom drawer of the cabinet is stuck. Can you give it a pull? \\
    \ttfamily [08:11:10] & Jeremy & Let me see... Push it in a bit, then give it a sharp tug – There, it's open. \\
    \ttfamily [08:11:18] & Jane & Wow, how did this box get so dusty? I think it's the old photo albums, right? \\
    \ttfamily [08:11:24] & Jeremy & Should be. That was before we switched to a digital camera, all these were developed from film. \\
    \ttfamily [08:11:30] & Jane & This one... was our first trip to Hangzhou? You were wearing that blue checkered shirt that year. \\
    \ttfamily [08:11:36] & Jeremy & Haha, yes, taken at the entrance of Lingyin Temple. You insisted that monk was peeking at us while we took the picture. \\
    \ttfamily [08:11:42] & Jane & He was looking! And you started laughing, the photo turned out all blurry. \\
    \ttfamily [08:11:50] & Jeremy & Check the back, I think there are some from that Yunnan trip too? \\
    \ttfamily [08:12:00] & Jane & Yes, here they are. At the gate of Dali Old Town, you had to wear your sunglasses crooked, trying to look all artsy. \\
    \ttfamily [08:12:06] & Jeremy & That was called `creating a vibe'. Look how happy you're laughing in this one. \\
    \ttfamily [08:12:12] & Jane & Hmm... My hair wasn't gray back then. \\
    \ttfamily [08:12:20] & Jeremy & It wasn't that long ago, was it? Seven or eight years? \\
    \ttfamily [08:12:25] & Jane & Almost. Time really flies. Oh, how did this USB drive box get here too? \\
    \ttfamily [08:12:32] & Jeremy & Used that for storing photos ages ago. I think it's labeled ``2016 Family Photos''. \\
    \ttfamily [08:12:38] & Jane & Can we still read it? Should we find a computer and try? \\
    \ttfamily [08:12:42] & Jeremy & I'll try it later on my study computer. The port should still be compatible. \\
    \ttfamily [08:12:50] & Jane & No rush, let's sort these albums first. The old ones go on this side, the newer ones over here. \\
    \ttfamily [08:13:00] & Jeremy & This yellow-edged one was from your mom, right? She said we should pick only the best ones to develop and keep. \\
    \ttfamily [08:13:06] & Jane & Yes, she kept saying back then that when we got old, we could look through them together. \\
    \ttfamily [08:13:12] & Jeremy & She was right. Isn't it nice looking through them now? \\
    \ttfamily [08:13:20] & Jane & Mmm... This box also has a group photo from Weizhou's wedding. \\
    \ttfamily [08:13:26] & Jeremy & Oh, look at him in the suit with a bow tie, like he stepped right out of a period drama set in the Republic of China era. \\
    \ttfamily [08:13:32] & Jane & You're one to talk! Your tie was crooked, and he had to retie it for you. \\
    \ttfamily [08:13:38] & Jeremy & Haha, you remember everything. We gotta keep this photo to tease him with next time we see him. \\
    \ttfamily [08:13:45] & Jane & Don't overdo it. He's ``Boss Zhang'' now, you know. \\
    \ttfamily [08:13:50] & Jeremy & To me, he'll always be that goofball who fell into the flowerbed playing basketball. \\
    \ttfamily [08:14:00] & Jane & Oh, this one is of your dad fixing his bike in the yard... \\
    \ttfamily [08:14:06] & Jeremy & Yeah, that old Phoenix brand bike. The chain kept falling off, he'd spend the whole afternoon fixing it. \\
    \ttfamily [08:14:12] & Jane & He was so handy. All your repair skills, you learned from him. \\
    \ttfamily [08:14:18] & Jeremy & Yeah... This is a really good photo. The light on his face, so peaceful. \\
    \ttfamily [08:14:25] & Jane & Let's not throw these old things away. Let's find a box and store them properly. \\
    \ttfamily [08:14:30] & Jeremy & Okay, I'll go get a storage bin from the storage room later. \\
    
\end{longtable}

\begin{longtable}{p{2cm} >{\bfseries}p{1cm} p{11cm}}
    \hiderowcolors
    \caption{Jeremy with Family Watching Spring Festival Gala (2024-02-10)} \label{tab:gala_sample} \\
    \toprule
    \textbf{Time} & \textbf{Speaker} & \textbf{Utterance} \\
    \midrule
    \endfirsthead
    
    \multicolumn{3}{c}{{\bfseries \tablename\ \thetable{} -- continued from previous page}} \\
    \toprule
    \textbf{Time} & \textbf{Speaker} & \textbf{Utterance} \\
    \midrule
    \endhead
    
    \bottomrule
    \multicolumn{3}{r}{{Continued on next page...}} \\
    \endfoot
    
    \bottomrule
    \endlastfoot
    
    \ttfamily [16:00:12] & Jeremy & Mom, Jane, the Spring Festival Gala replay has started. The tea is freshly brewed, have it while it's hot. \\
    \ttfamily [16:00:18] & Mother & Oh, this tea aroma is so comforting. Hangzhou Longjing really is something else. \\
    \ttfamily [16:00:25] & Jane & Mmm, so light and refreshing. One sip and I feel completely relaxed. \\
    \ttfamily [16:03:40] & Mother & These hosts look the same as always, wearing red dresses every year, smiling like flowers. \\
    \ttfamily [16:05:10] & Jeremy & Mom, don't just look at what they're wearing. There's a cross-talk performance later, you love those. \\
    \ttfamily [16:08:33] & Jane & I recognize this skit actor. He was hilarious last time playing that delivery guy. \\
    \ttfamily [16:12:15] & Mother & Oh my, this kid acts so well, the way he talks is exactly like Auntie Wang next door back in my hometown. \\
    \ttfamily [16:18:44] & Jeremy & Here, Mom, let me top up your tea. Careful, don't spill. \\
    \ttfamily [16:19:01] & Jane & Did Dad used to love watching the Gala too? I remember you saying he always liked memorizing the punchlines. \\
    \ttfamily [16:19:10] & Mother & Oh yes, your father-in-law would even take notes in a little notebook, saying he'd tell the students when school started. \\
    \ttfamily [16:25:20] & Jeremy & This cross-talk is okay, but not as good as last year's. \\
    \ttfamily [16:27:05] & Jane & Don't be so picky. Just being able to sit and watch it together as a family is nice enough. \\
    \ttfamily [16:30:18] & Mother & Oh, speaking of cross-talk, it just reminded me—when Mingyuan was little, he went to pick bayberries on the hill behind the village. He fell out of the tree but insisted he didn't! \\
    \ttfamily [16:30:30] & Jeremy & Mom, not this story again... \\
    \ttfamily [16:30:33] & Jane & Huh? Tell me, tell me! I haven't heard this one! \\
    \ttfamily [16:30:38] & Mother & That day, he insisted the sweetest bayberries were on the highest branch. Well, his hand slipped, and he landed right on his backside. Came back still stubbornly saying ``I didn't cry,'' but his face was all swollen. Saying that with one side of his face puffed up, he looked like a little steamed bun. \\
    \ttfamily [16:31:05] & Jane & Huh? Stung by a bee? Did you just say a bee? \\
    \ttfamily [16:31:08] & Mother & Oh yes, right, it was a bee! I got mixed up—that was another time! Picking wild strawberries, there was a beehive in the grass, ``buzz'' and it stung him right on the face! \\
    \ttfamily [16:31:18] & Jeremy & I really didn't cry, it's just... the tears came out on their own. \\
    \ttfamily [16:31:22] & Jane & Hahaha, stop it! ``Tears came out on their own''? What's that if not crying? \\
    \ttfamily [16:31:27] & Mother & Exactly! He was so swollen even your dad couldn't recognize him, still insisting ``I didn't cry.'' I put a cold towel on his face, and he's sniffling, saying ``It's just a little itchy.'' \\
    \ttfamily [16:31:40] & Jane & That's adorable! I have to write this down—(sound of typing on phone) Title it ``Future Parenting Material''. \\
    \ttfamily [16:31:48] & Jeremy & Hey, don't write that down. What kind of positive example is that... \\
    \ttfamily [16:31:52] & Mother & Why not? Stubborn kid, full of spirit! Kids these days don't have that kind of grit anymore. \\
    \ttfamily [16:32:10] & Jane & When we... if we have kids in the future, I'll tell them this story. I'll add a subtitle: ``On the Art of Graceful Stubbornness''. \\
    \ttfamily [16:32:18] & Jeremy & Don't you two gang up on me... \\
    \ttfamily [16:32:25] & Mother & This isn't ganging up, it's family memories! Come on, Mingyuan, pour some more tea, let's keep watching. \\
    \ttfamily [16:35:40] & Jane & This dance is so beautiful, the backdrop looks like an ink wash painting. \\
    \ttfamily [16:36:15] & Mother & Yes, the costumes are lovely too, the colors are elegant, not too flashy. \\
    \ttfamily [16:40:30] & Jeremy & The special effects here are used quite cleverly, they sync up well with the performers' movements. \\
    \ttfamily [16:42:10] & Jane & See, isn't this what you called ``cross-boundary integration''? \\
    \ttfamily [16:42:15] & Jeremy & Heh, I guessed the start, but I didn't expect the effects to be this smooth. \\
    \ttfamily [17:00:20] & Mother & This song is sung so beautifully, warms your heart listening to it. \\
    \ttfamily [17:05:35] & Jane & This skit is starting to get interesting. This dad acts exactly like the department head at my clinic. \\
    \ttfamily [17:10:12] & Jeremy & Shh—the accompaniment is coming up, I really like this melody. \\
    \ttfamily [17:30:45] & Mother & Oh my, it's almost six o'clock. Shouldn't we start preparing dinner? \\
    \ttfamily [17:31:00] & Jeremy & No rush. I've got some chicken soup with Chinese yam simmering, just need to heat it up, and there are dumplings too. \\
    \ttfamily [17:31:10] & Jane & I'll set the table and pan-fry the leek dumplings we made yesterday. \\
    \ttfamily [17:31:18] & Mother & Good, I'll help you with the tea. Time just flies when you're drinking this tea. \\
    \ttfamily [19:02:10] & Jane & The song and dance numbers on the Gala are one after another, it's making me sleepy. \\
    \ttfamily [19:02:25] & Mother & Yes, when I was young I could stay up until midnight, but now I feel like closing my eyes past nine. \\
    \ttfamily [19:03:05] & Jeremy & How about we take a break? We can get up again closer to midnight? \\
    \ttfamily [19:03:12] & Jane & Okay, I'll go charge my phone first, and I need to organize my notes. \\
    \ttfamily [19:03:20] & Mother & I'll just stay put here. You two go ahead, I'll just listen to the Gala. \\
\end{longtable}

\begin{longtable}{p{2cm} >{\bfseries}p{1cm} p{11cm}}
    \hiderowcolors
    
    \caption{Jeremy in Emergency Project Post-Mortem Meeting (2024-06-03)} \label{tab:meeting_sample} \\
    \toprule
    \textbf{Time} & \textbf{Speaker} & \textbf{Utterance} \\
    \midrule
    \endfirsthead
    
    \multicolumn{3}{c}{{\bfseries \tablename\ \thetable{} -- continued from previous page}} \\
    \toprule
    \textbf{Time} & \textbf{Speaker} & \textbf{Utterance} \\
    \midrule
    \endhead
    
    \bottomrule
    \multicolumn{3}{r}{{Continued on next page...}} \\
    \endfoot
    
    \bottomrule
    \endlastfoot
    
    \ttfamily [10:15:00] & Jeremy & Is everyone here? Let's get started. As you all saw, last night's incident had a significant impact. We need to quickly piece together the timeline and identify the root causes. \\
    \ttfamily [10:15:45] & Mike & Yes, we came straight from the morning stand-up. Wei and the Ops reps are here too. \\
    \ttfamily [10:15:55] & Jeremy & Good. Let me briefly recap the timeline. Last night at 21:47, our monitoring platform started receiving a flood of 503 errors, concentrated on the user login and permission verification APIs. Frontend service response times spiked from an average of 80 milliseconds to over two seconds, lasting roughly twenty minutes. \\
    \ttfamily [10:17:10] & Alex & On the backend side, we didn't receive alerts until 21:49, two minutes after the problem started. Furthermore, the initial alerts were scattered; no one realized it was a systemic issue initially. \\
    \ttfamily [10:17:45] & Wei & The test environment monitoring didn't trigger because we hadn't simulated failure states for that authentication component. It appears a vulnerability in the third-party SDK was triggered by a scanning tool, causing it to crash outright, which then cascaded to our authorization service. \\
    \ttfamily [10:18:35] & Other & Correct. Checking the logs confirms it's the CVE-2024-3187 mentioned in their urgent patch bulletin – a high-severity privilege escalation vulnerability. When their service restarted, our persistent connections were all severed, and we lacked reconnection safeguards. \\
    \ttfamily [10:19:40] & Jeremy & So, fundamentally, it wasn't our code at fault. But the core issue is that our monitoring didn't flag the anomaly immediately. From 21:47 to 21:58 – a full 11 minutes – there was no clear, high-severity ``service meltdown'' alert. \\
    \ttfamily [10:20:35] & Mike & That's unacceptable. Users couldn't access the app, and we were in the dark? \\
    \ttfamily [10:20:50] & Jeremy & Exactly. Reviewing the Grafana dashboards, while we had heartbeat metrics, we lacked aggregated alerting for them. Also, the alert rules are too fragmented; a sea of red dots ended up masking the critical issue. \\
    \ttfamily [10:21:45] & Wei & I checked the logs last night. The first call was to Alex at 21:55, reporting login timeouts. That's when we first suspected a common problem, but the command chain was unclear – no one took clear ownership of the emergency response. \\
    \ttfamily [10:22:30] & Alex & I was initially checking logs, thought it might be a database issue, and even had the DBA team investigate. It took time to realize the upstream auth service was the root cause. \\
    \ttfamily [10:23:15] & Ops Rep & We were also reactive. By the time we noticed the abnormal traffic drop and intervened, the golden window for mitigation had passed. \\
    \ttfamily [10:24:00] & Jeremy & Therefore, while the trigger was a third-party component failure, this incident exposed our own weaknesses: insufficient monitoring sensitivity and a lack of a formalized emergency response process. \\
    \ttfamily [10:24:50] & Mike & Agreed. The responsibility for the cause isn't ours, but our response was too slow. This has to change. \\
    \ttfamily [10:25:15] & Jeremy & I propose we focus on two key areas moving forward. First, integrate health checks for external dependencies into our core monitoring. Heartbeat, version status, abnormal reconnection states – all need real-time, prominent alerting. \\
    \ttfamily [10:26:20] & Wei & We can integrate that with our existing component health dashboard. Wasn't that already in progress? \\
    \ttfamily [10:26:40] & Jeremy & Yes, this fits perfectly. Second, I've been thinking since last night: we need to prioritize implementing a robust canary release and automated rollback mechanism. If we could have automatically detected the spike in abnormal call rates and rolled back to the previous stable version, we could have halved the outage duration. \\
    \ttfamily [10:27:55] & Alex & Automated rollback? Isn't that a bit aggressive? What about false positives? \\
    \ttfamily [10:28:20] & Jeremy & Not a full, automatic rollback for all traffic. We can start with a canary release for a small percentage of users, say 1\%, while closely monitoring key metrics – error rate, latency, authentication failure rate. If these exceed thresholds, automatically route traffic back to the old version and trigger alerts. \\
    \ttfamily [10:29:30] & Ops Rep & We support this approach. We can configure the traffic switching using K8s; we've tested similar setups in our test environment before. \\
    \ttfamily [10:30:10] & Wei & Then our release process needs updating too. The current manual tagging and manual image push is prone to missed steps. \\
    \ttfamily [10:30:45] & Jeremy & Exactly. I want to implement a pre-release checklist, similar to the one we drafted earlier. Items like dependency scans, permission verification, rollback plan confirmation – all must be checked off before deployment. \\
    \ttfamily [10:31:40] & Mike & I agree with this direction. Especially regarding external dependencies, we must confirm there are no known vulnerabilities and have a degradation plan before any future deployment. \\
    \ttfamily [10:32:25] & Jeremy & I'll take the lead on drafting an improvement plan covering monitoring enhancements, the release process, and the emergency response mechanism. Target is to have a first draft by the end of this week. \\
    \ttfamily [10:33:15] & Mike & Okay. You coordinate. Wei, you support with testing validation. Ops team, please provide a feasibility report for the automated traffic switching. \\
    \ttfamily [10:33:55] & Ops Rep & Understood. We can schedule a technical alignment meeting this afternoon. \\
    \ttfamily [10:34:25] & Jeremy & Good. Additionally, I suggest we conduct a failure drill next week, simulating a third-party service outage, to test if our current response procedures can handle it. \\
    \ttfamily [10:35:15] & Wei & Agreed. I'll design the scenario, maybe add some complications like alerts being incorrectly marked as low priority. \\
    \ttfamily [10:36:00] & Alex & I'll prepare an emergency procedure document then, clarifying roles and responsibilities – who does what under which circumstances – to prevent the lack of leadership we saw. \\
    \ttfamily [10:36:50] & Jeremy & Alright, let's proceed on that basis. We'll schedule follow-up meetings for the details. Let's wrap up this post-mortem for now? \\

\end{longtable}

\clearpage

\subsection{Question Types and Examples}
\label{app:question_types_examples}

To provide a clearer understanding of the evaluation tasks, we first present the formal definitions of the four distinct question types designed in \textsc{LifeDialBench}, followed by concrete examples from the dataset.

\begin{itemize}[leftmargin=*]
    \item \textbf{QT1: Event Content Recall.} This type encompasses questions that demand the retrieval of core event content, and it falls under the broader category of Event Recall.
    
    \item \textbf{QT2: Event Detail Retrieval.} Questions of this type require precise retrieval of specific event details, and they are classified under Detail Retrieval.
    
    \item \textbf{QT3: Multi-hop Event Reasoning.} These questions involve both retrieving and reasoning across multiple events, and they belong to the Temporal Reasoning category.
    
    \item \textbf{QT4: Temporal Grounding.} As a lifelog-specific subcategory of Detail Retrieval, this unique type requires accurately pinpointing the exact timestamp of a particular event to generate a valid answer.
\end{itemize}

\vspace{0.5em}
\noindent The following examples demonstrate these question types, featuring the query, associated timestamp, and candidate options.

\begin{benchbox}[Question Examples]
    \setlength{\parskip}{0.8em} 

    \qtype{Single Event:} What was the main topic of discussion between Jeremy and Jane during the organization of old items? 
    \timestamp{query\_timestamp=2024-01-03}
    \begin{itemize}[nosep, leftmargin=1.5em, label=$\bullet$]
        \item \textbf{(A)} Memories of their 2018 trip to Dali and Lijiang in Yunnan; 
        \item \textbf{(B)} Preliminary planning for the Spring Festival holiday; 
        \item \textbf{(C)} Optimization solutions for household clutter management; 
        \item \textbf{(D)} Discussion on edge computing communication protocols.
    \end{itemize}
    
    \lightsep

    \qtype{Event Detail:} What specific item did Jane mention when recalling the Yunnan trip? 
    \timestamp{query\_timestamp=2024-01-01}
    \begin{itemize}[nosep, leftmargin=1.5em, label=$\bullet$]
        \item \textbf{(A)} A tie-dyed scarf; 
        \item \textbf{(B)} A bicycle; 
        \item \textbf{(C)} A hat; 
        \item \textbf{(D)} A pair of shoes.
    \end{itemize}
    
    \lightsep

    \qtype{Multi Event:} During which activities did Jeremy and Jane discuss topics related to children's health? 
    \timestamp{query\_timestamp=2024-01-01}
    \begin{itemize}[nosep, leftmargin=1.5em, label=$\bullet$]
        \item \textbf{(A)} During breakfast and balcony reading; \item \textbf{(B)} While organizing old items and watching a movie; 
        \item \textbf{(C)} During grocery shopping and dinner preparation; 
        \item \textbf{(D)} During lunch and while debugging the projector.
    \end{itemize}
    
    \lightsep

    \qtype{Temporal Info:} What was the specific time when Jeremy and Jane began immersing themselves in the photos from their Yunnan trip? 
    \timestamp{query\_timestamp=2024-01-03}
    \begin{itemize}[nosep, leftmargin=1.5em, label=$\bullet$]
        \item \textbf(A) 9:00 AM; \item \textbf{(B)} 10:30 AM; \item \textbf{(C)} 11:00 AM; \item \textbf{(D)} 1:00 PM.
    \end{itemize}

\end{benchbox}

\section{Prompts}
\label{app:prompts}
\subsection{Judge Prompt used for Open-Ended Format}
\label{Judge Prompt}
To evaluate the semantic accuracy of the Open-Ended generation, we employ an LLM-based judge. This judge compares the model's response against the ground-truth answer (derived from the correct option) to determine semantic equivalence. The specific prompt used is detailed below:

\begin{benchbox}[Judge Prompt for Open-Ended QA]
You are given a question, its ground-truth answer, and a model response. Determine if the model response is semantically equivalent or meaningfully similar to the ground-truth answer. Consider the following as acceptable variations:
\begin{itemize}
    \item Different wording but same core meaning
    \item Partial answers that contain the key information
    \item Answers with additional relevant context
    \item Answers that rephrase the same idea
    \item Minor factual details may differ if the main point is correct
\end{itemize}

Be lenient in your judgment - if the response captures the essence of the correct answer, consider it correct.

\vspace{0.5em}
\noindent Question: \{question\} \\
Ground-truth answer: \{reference\} \\
Model response: \{candidate\}
\end{benchbox}

\subsection{Lifelog Generation Prompt}
Here is the prompt we use for transforming 10-minutes summarization to lifelog.
\begin{benchbox}[Prompt Template]
You are required to transform the target first-person narrative into lifelog-style conversation records. **Lifelog** refers to authentic daily spoken conversations captured by portable recording devices. 
Your task is not storytelling but converting the given narrative into natural dialogues that sound like real speech.\vspace{10pt}\\
\# Character name\\
\{character\_name\}\vspace{10pt}\\
\# Previous Narratives (context for coherence):\\
\{previous\_narratives\}\vspace{10pt}\\
\# Target First-person Narrative: \\
\{first\_person\_narrative\}\vspace{10pt}\\
\# Time range in target narrative:\\
\{time\_range\}\vspace{10pt}\\
\# Conversation Generation Requirements\\
**Core Conversion Principles:**\\
1. **Narrative-to-Lifelog Transformation**: Convert the target first-person narrative into lifelog dialogues, ensuring all important details from the narrative are preserved in the conversations.  \\
2. **Continuity and Non-redundancy**: Previous narratives are provided to maintain timeline consistency, character relationships, and avoid repeating the same details unnecessarily.  \\
3. **Authenticity**: The dialogues must sound natural, spontaneous, and spoken in real daily English, avoiding formal or literary expressions.  \vspace{10pt}\\
**Format Specifications:**\\
- Strictly use the format:  
  [yyyy-mm-dd, HH:MM:SS] Character: Speech content  \vspace{10pt}\\
**Content Requirements:**\\\
1. **Detail Preservation**: Every concrete detail in the target narrative (actions, observations, emotions, objects, times, etc.) must appear in the dialogues.  \\
2. **Logical Flow**: Keep the event flow consistent with both the target narrative and previous lifelogs.  \\
   - Ensure continuity of relationships between characters.  \\
   - Keep the timeline reasonable and coherent.  \\
3. **Boundary Control**: Do not introduce cross-day planning, greetings, farewells, or artificial summaries. End conversations naturally when the described event ends.  \vspace{10pt}\\
**Output Format:**\\
- Only output lifelog dialogues in English, without explanations, notes, or extra text. \vspace{10pt}\\ 
\# Example Format\\
\text{[2025-09-17, 09:23:11]} Speaker A: Actual spoken words  \\
\text{[2025-09-17, 09:23:15]} Speaker B: Dialogue continues  \vspace{10pt}\\

Now please generate lifelog conversations according to the above requirements.
\end{benchbox}

The following prompts were employed in the Top-Down Hierarchical Life Simulation Framework. Year-level summaries are progressively allocated and enriched at the month level to generate detailed monthly summaries, while the prompts for the "month-to-week" and "week-to-day" stages have been slightly adjusted.

\begin{benchbox}[Prompt Template to Allocate]
You are a professional lifelog analyst. Based on the provided annual experience summary, restructure and expand the content by month to generate detailed, coherent, and realistic monthly life records.\vspace{10pt}\\
\{holidays\}\\
\{important\_days\}\vspace{10pt}\\
\# Annual Experience Summary:\\
\{year\_summary\}\vspace{10pt}\\
\# Requirements\\
- Each monthly record must clearly describe the time, location, people involved, process, and outcomes of events.\\
- While strictly reconstructing the annual experiences by month, you may expand each month's record.\\
- Your expansions must be realistic; ensure the content is substantial and natural, and avoid fabricated dramatic plots or supernatural elements.\\
- After reconstruction and expansion, each month's record must cover major events, work, exercise, entertainment, family communication, and social activities.\\
- If a specific time point for an event is clearly stated in the annual summary, you must not change it; if it is not specified, assign a reasonable time.\vspace{10pt}\\
\# Output Format\\
Output strictly as a standard JSON array, and output only the JSON array without any explanations or comments. Each item in the JSON array should have the following structure:\\
\text{[}
    \{\{
        "Month": "\{year\} January",
        "Monthly Record": "..."
    \}\},
    \{\{
        "Month": "\{year\} February",
        "Monthly Record": "..."
    \}\},
    ...
\text{]}
\end{benchbox}

\begin{benchbox}[Prompt Template to Enrich]
You are a professional lifelog analyst. Below are this person's monthly records for the target month and the adjacent months. Please enrich the current record for the target month to make the description more comprehensive.\vspace{10pt}\\
\{prev\_months\}\vspace{10pt}\\
\# Existing monthly record for \{month\}:\\
\{month\_data\}\vspace{10pt}\\
\# Date information for \{month\}:\\
\{month\_dates\_info\}\vspace{10pt}\\
\# Requirements:\\
- Each monthly record must clearly describe the time, location, people involved, process, and outcomes of events.\\
- Unless the current month's record already contains such mentions, do not add any cross-month plans during enrichment; for example, do not schedule April activities in the March record.\\
- The enriched content must be realistic; ensure the content is substantial and natural. Avoid fabricated dramatic plots or supernatural elements.\\
- The enriched record should cover all facets of life, including but not limited to major events, work, exercise, entertainment, family communication, and social activities.\\
- The enriched content must cover the entire month—early, mid, and late—and distribute events as evenly as possible. If the original record provides specific dates/times, you must keep them.\\
- The enriched content must remain temporally consistent with the records of the previous and following months, ensuring coherence without contradictions.\\
- Note that workdays are typically Monday through Friday, rest days are Saturday and Sunday, and public holidays are rest days. Arrange work and life content accordingly.\vspace{10pt}\\
\# Output Format\\
Output strictly as standard JSON, and output only the JSON without any explanations or comments. The JSON fields are:\\
\{
  "Month": "\{month\}",
  "Monthly Record": "..."
\}
\end{benchbox}

\subsection{Question Generation Prompt}
\label{app:question generation prompt}
\begin{benchbox}[Prompt Template to Generate Daily-level Questions]
\# Prompt for Event Extractor Evaluation Data Generation\\
You need to generate evaluation data for an event extractor. The event extractor will extract useful information from users' life records and store it in a database.  \\
Now you will be provided with a user's daily experiences, and you need to generate four questions based on the content, with four options (A, B, C, D) for each question (one correct answer and three distractor options). These questions and options will be used to evaluate the extraction performance of the event extractor.\\

\#\# Daily Events ({date})\\
\{all\_day\_events\}\\

\#\# Question Requirements\\
- Generate 4 question-answer pairs, which should ask about the following four aspects respectively:\\
  - The content of a specific event\\
  - A specific detail of a specific event\\
  - The content of multiple events\\
  - The specific time when a specific event occurred\\
- Question Guidelines:\\
  - Frame questions about events that involve interactions with others and can generate dialogue data; do not frame questions about events that cannot generate dialogue data.\\
  - The events targeted by the questions must be unique enough and must not be daily routine events.\\

\#\# Output Requirements\\
- You need to output a JSON list, where each JSON element contains the following fields:\\
  - `question`: The content of the question\\
  - `options`: A list containing four options, formatted as ["A. Option content", "B. Option content", "C. Option content", "D. Option content"]\\
  - `answer`: The option letter of the correct answer, e.g., `A`\\
- Do not output any content other than the JSON list of question-answer pairs
\end{benchbox}

\section{Additional Results}
\subsection{Offline Evaluation Results}
\noindent We present the performance of various memory systems in the \textbf{offline evaluation setting} (\Cref{tab:offline-results}). In this configuration, agents process the complete dialogue lifelog before answering queries. These results provide a benchmark for the models' fundamental memory capacity, offering a performance upper bound by decoupling the memory task from the requirements of real-time, causal streaming.
\begin{table}[h]
\centering
\setlength{\tabcolsep}{4.5pt} 
\renewcommand{\arraystretch}{1.2} 
\caption{Results on Offline Evaluation Setting}
\label{tab:offline-results}
\resizebox{\linewidth}{!}{
\begin{tabular}{c|c|ccccc|ccccc}
\toprule
 & & \multicolumn{5}{c|}{Open-Ended} & \multicolumn{5}{c}{Multiple-Choices} \\
Models & Method & QT1 & QT2 & QT3 & QT4 & Overall & QT1 & QT2 & QT3 & QT4 & Overall \\
\midrule
\multicolumn{12}{c}{\textbf{EgoMem}} \\
\midrule
\multirow{4}{*}{gpt-4o-mini} 
 & RAG   & 37.09 & 51.25 & 9.60 & 23.87 & 30.88 & 68.14 & 83.75 & 56.33 & 49.09 & 64.74 \\
 & A-Mem & 33.06 & 51.25 & 9.17 & 27.47 & 30.56 & 68.54 & 81.25 & 55.02 & 54.05 & 64.74 \\
 & Mem0  & 12.09 & 16.25 & 2.62 & 14.41 & 11.39 & 54.03 & 51.35 & 46.28 & 32.88 & 46.11 \\
 & MemOS & 26.20 & 32.50 & 9.60 & 9.90  & 19.91 & 66.93 & 72.08 & 68.55 & 32.88 & 60.59 \\
\midrule
\multirow{4}{*}{qwen-plus} 
 & RAG   & 38.70 & 56.25 & 11.79 & 32.43 & 35.14 & 66.12 & 82.91 & 55.89 & 51.80 & 64.53 \\
 & A-Mem & 38.30 & 52.91 & 13.10 & 29.72 & 33.86 & 67.33 & 80.41 & 51.09 & 55.40 & 63.89 \\
 & Mem0  & 12.90 & 17.08 & 3.93  & 10.36 & 11.18 & 40.72 & 51.25 & 34.93 & 33.33 & 40.25 \\
 & MemOS & 29.83 & 42.08 & 13.53 & 31.53 & 29.39 & 62.50 & 79.58 & 61.13 & 52.25 & 64.11 \\
\midrule
\multicolumn{12}{c}{\textbf{LifeMem}} \\
\midrule
\multirow{4}{*}{gpt-4o-mini} 
 & RAG   & 32.41 & 71.80 & 19.99 & 20.18 & 36.12 & 73.79 & 91.70 & 69.07 & 52.67 & 72.13 \\
 & A-Mem & 28.27 & 65.17 & 17.20 & 19.71 & 32.61 & 71.95 & 88.85 & 68.36 & 46.16 & 69.14 \\
 & Mem0  & 5.97  & 21.56 & 4.19  & 5.80  & 9.36  & 66.89 & 62.56 & 67.67 & 31.09 & 57.37 \\
 & MemOS & 23.21 & 55.17 & 11.26 & 13.79 & 25.85 & 73.56 & 78.61 & 66.89 & 30.80 & 62.47 \\
\midrule
\multirow{4}{*}{qwen-plus} 
 & RAG   & 35.17 & 68.24 & 25.81 & 31.32 & 40.22 & 74.02 & 90.52 & 69.76 & 57.20 & 72.80 \\
 & A-Mem & 34.70 & 68.71 & 29.99 & 29.70 & 40.86 & 72.64 & 87.44 & 62.09 & 54.88 & 69.19 \\
 & Mem0  & 11.49 & 30.33 & 8.83  & 9.97  & 15.16 & 68.50 & 68.72 & 58.37 & 40.23 & 58.94 \\
 & MemOS & 36.36 & 69.31 & 25.45 & 20.45 & 37.89 & 73.18 & 83.86 & 65.68 & 56.59 & 69.82 \\
\bottomrule
\end{tabular}
}
\end{table}

\subsection{Question Types Distribution}\label{app:questino types distribution}
\begin{figure}[htbp]
    \centering 
    \includegraphics[width=0.5\linewidth]{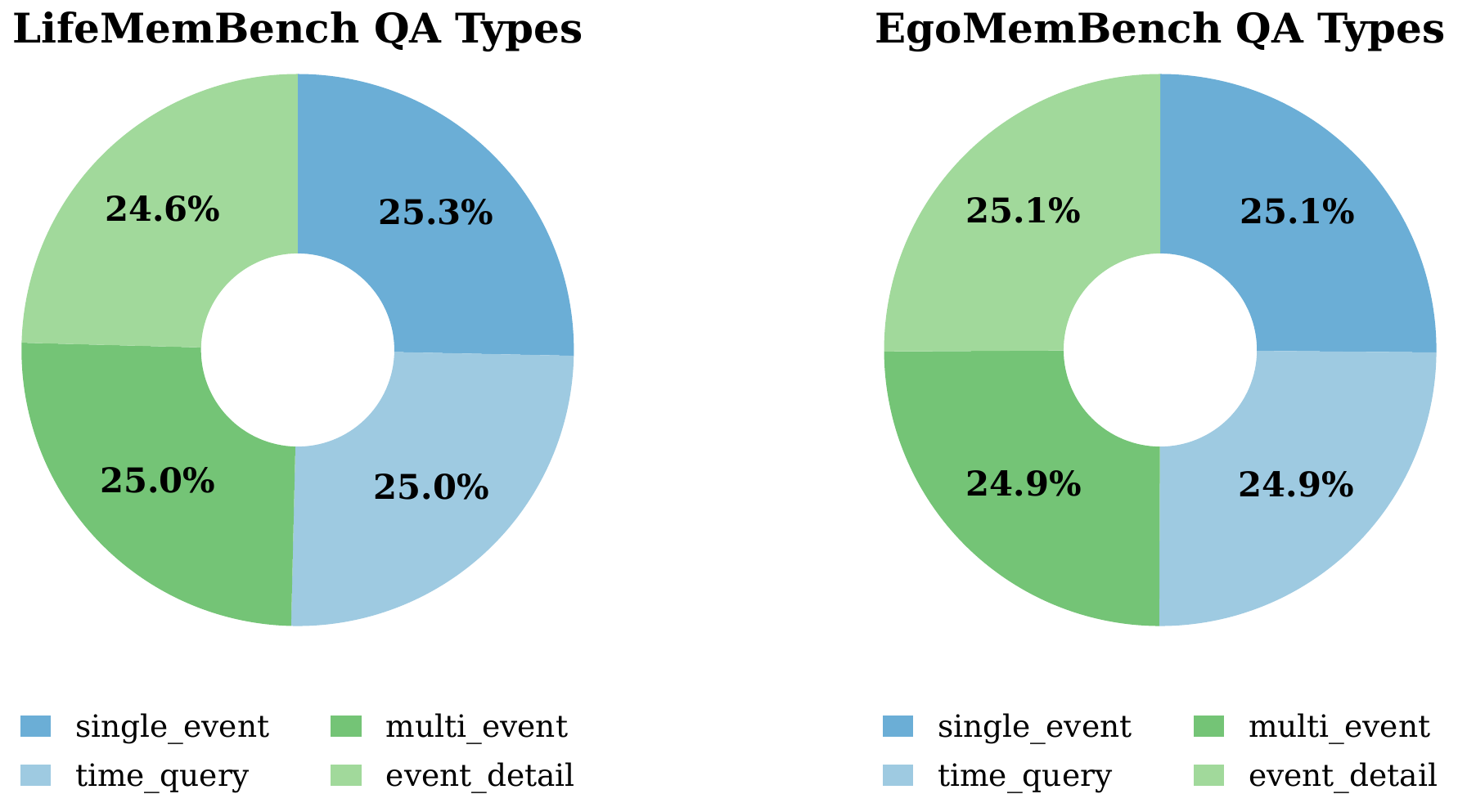}
    \caption{The Distribution of QA types.}
    \label{fig:category distribution}
\end{figure}

The distribution of four question types is in \Cref{fig:category distribution}. We ensure that both subsets would have a balance question-type proportion.

\section{Memory Systems}
\label{app:memory_system}
\subsection{Description of Memory Systems}
\begin{figure}[htbp]
    \centering
    \includegraphics[width=\linewidth]{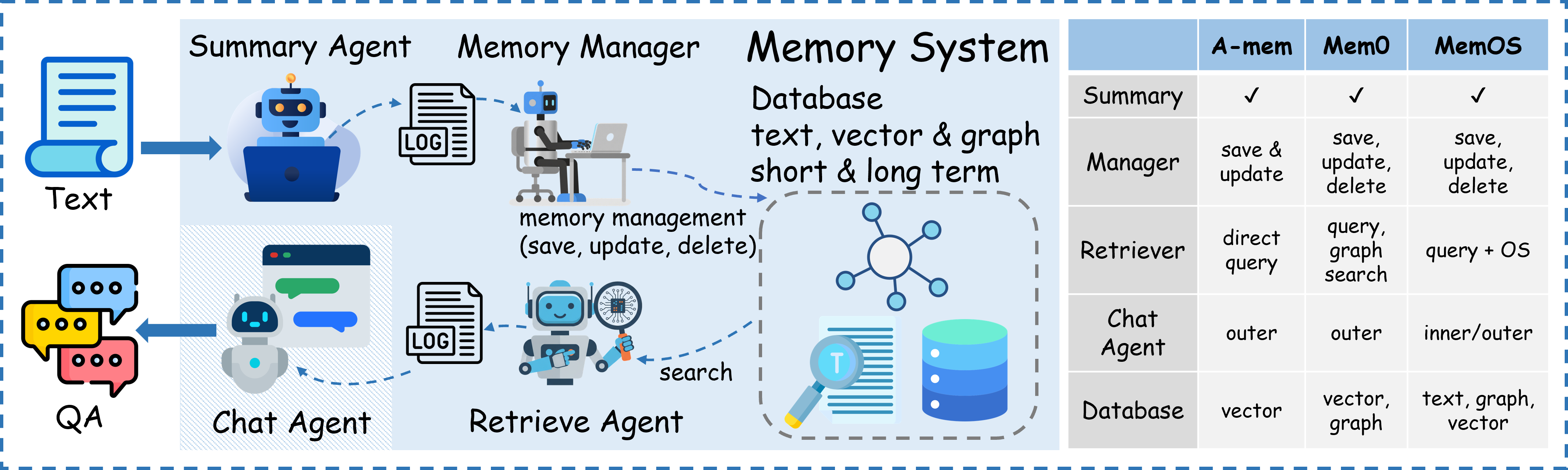}
    \caption{Definition of the structure of a memory system, and a comparison table of current memory system approaches under this structure.}
    \label{fig:memory_system}
\end{figure}

We define the agent's characters in a memory system as \Cref{fig:memory_system}. A memory system often constructed by a summary agent, a memory manager, a retrieve agent, and a chat agent. The actual role would be different corresponding to the system design, as some role could be merged to one (e.g., memory manager and retrieve agent). The chat agent could be a part of the memory system, while there are also some systems exclude it.

\subsection{Implementation Details}
\label{app:implementation}
\paragraph{General Settings}
To ensure a fair comparison across all baselines, we maintain a unified configuration for retrieval and processing granularity. Specifically, we set the retrieval depth \textbf{top-$k$ to 20} by default. Furthermore, we adopt the \textbf{session (event)} as the fundamental processing unit for memory ingestion and storage.

\paragraph{RAG}
The simple RAG baseline includes a chat-agent and an embedding model to save and retrieve the relevant text. Therefore, there is no summary agent, no LLM memory manager, and no LLM retriever inside the system. It directly embeds and retrieves the lifelog text chunks into a vector database.

\paragraph{A-Mem, Mem0 and MemOS}
We follow the official code of these memory systems' GitHub repositories for evaluation. The prompts inside these systems are specifically refined to fit the requirement for our benchmark evaluation.

\end{document}